\documentclass[journal]{IEEEtran}

\usepackage{amsfonts}
\usepackage{epsfig}
\usepackage{psfrag}

\usepackage{relsize}
\usepackage[usenames, dvipsnames]{color}
\usepackage[font=small,skip=0pt]{caption}
\usepackage{leftidx}
\usepackage{microtype}

\usepackage{amsthm}
\newtheorem{definition}{Definition}
\newtheorem{lemma}{Lemma}

\usepackage{enumitem}
\usepackage{amsmath}
\usepackage{cleveref}
\usepackage{amssymb}
\usepackage{subfigure}
\usepackage[table,xcdraw]{xcolor}
\usepackage{graphicx}
\usepackage{chemmacros}
\usepackage{siunitx}
\usepackage[ruled,vlined,lined,ruled,linesnumbered]{algorithm2e}
\usepackage{latexsym,amsthm,array,amsfonts,booktabs,graphicx,subfigure,multirow,cuted,stfloats}
\usepackage{footmisc}
\usepackage{cite}
\usepackage{tabularx}
\usepackage{bm}
\usepackage{float}
\usepackage{epstopdf}
\usepackage{threeparttable}
\usepackage{multirow,booktabs,color,soul,threeparttable}

\definecolor{hl}{rgb}{0.75,0.75,0.75}
\sethlcolor{hl}
\newlength\savewidth

\SetCommentSty{mycommfont}

\begin{document}
\title{A Secure Federated Data-Driven Evolutionary Multi-objective Optimization Algorithm 
}
\author{Qiqi~Liu\textsuperscript{*},
         Yuping Yan\textsuperscript{*},
         P\'eter Ligeti,
         and~Yaochu~Jin,~\IEEEmembership{Fellow,~IEEE}

 \thanks{©2023 IEEE. Personal use of this material is permitted.  Permission from IEEE must be obtained for all other uses, in any current or future media, including reprinting/republishing this material for advertising or promotional purposes, creating new collective works, for resale or redistribution to servers or lists, or reuse of any copyrighted component of this work in other works.}
 \thanks{ This research is funded by the National Natural Science Foundation of China under Grant No. 62302147, China. Yaochu Jin is funded by an Alexander von Humboldt Professorship for Artificial Intelligence endowed by the Federal Ministry of Education and Research, Germany.  This research was partially supported by the project No. 2019-1.3.1-KK-2019-00011 financed by the National Research, Development and Innovation Fund of Hungary under the Establishment of Competence Centers, Development of Research Infrastructure Programme funding scheme and by ERC Advanced Grant “ERMiD".}
 \thanks{Qiqi Liu is with Trustworthy and General AI Lab, School of Engineering, Westlake University, China, and with Faculty of Technology, Bielefeld University, 33619 Bielefeld, Germany. e-mail: qiqi6770304@gmail.com (Qiqi Liu and Yuping Yan contribute equally).}
  \thanks{Yuping Yan is with the Faculty of Informatics, Department of Computeralgebra, Eötvös Loránd University, Hungary. She is also with Smart Data Group, E-Group ICT Software Zrt, Hungary. e-mail: yupingyan@inf.elte.hu}
 \thanks{P\'eter Ligeti is with the Faculty of Informatics, Department of Computeralgebra, Eötvös Loránd University, Hungary. e-mail: ligetipeter@inf.elte.hu}
 \thanks{Yaochu Jin is with the Faculty of Technology, Bielefeld University, 33619 Bielefeld, Germany. He is also with the Department of Computer Science, University of Surrey, Guildford, Surrey GU2 7XH, UK. email: yaochu.jin@uni-bielefeld.de. }
 \thanks{Corresponding Author: Yaochu Jin (yaochu.jin@uni-bielefeld.de)}

 }

\maketitle

\begin{abstract}
Data-driven evolutionary algorithms usually aim to exploit the information behind a limited amount of data to perform optimization, which have proved to be successful in solving many complex real-world optimization problems. However, most data-driven evolutionary algorithms are centralized, causing privacy and security concerns. Existing federated Bayesian optimization algorithms and data-driven evolutionary algorithms mainly protect the raw data on each client. To address this issue, this paper proposes a secure federated data-driven evolutionary multi-objective optimization algorithm to protect both the raw data and the newly infilled solutions obtained by optimizing the acquisition function conducted on the server. We select the query points on a randomly selected client at each round of surrogate update by calculating the acquisition function values of the unobserved points on this client, thereby reducing the risk of leaking the information about the solution to be sampled. In addition, since the predicted objective values of each client may contain sensitive information, we mask the objective values with Diffie-Hellman-based noise, and then send only the masked objective values of other clients to the selected client via the server. Since the calculation of the acquisition function also requires both the predicted objective value and the uncertainty of the prediction, the predicted mean objective and uncertainty are normalized to reduce the influence of noise. Experimental results on a set of widely used multi-objective optimization benchmarks show that the proposed algorithm can protect privacy and enhance security with only negligible sacrifice in the performance of federated data-driven evolutionary optimization. 

\end{abstract}


\section{Introduction}

Evolutionary algorithms (EAs) are a class of population-based metaheuristics inspired from natural evolution that have witnessed huge success in solving single- or multi-objective optimization problems \cite{deb2002computationally}. Since they do not rely on gradients of the objective functions, EAs are well suited for data-driven surrogate-assisted optimization \cite{jin2021book}, often in combination with Bayesian optimization (BO) \cite{shahriari2016taking,wang2022survey} for surrogate management.   

Despite their effectiveness and efficiency, existing BO and data-driven EAs suffer from several limitations. {\color{black}{At present, most data-driven EAs are implemented in a centralized manner, assuming that all required information for optimization is available on a single computer, including raw data, the acquisition function, and new data points. However, this centralized approach is unpractical in many real-world scenarios, since raw data and data to be sampled may be distributed across multiple locations. For instance, as described in the scenario in \cite{kharkovskii2020private}, a bank (i.e., a client) aims to identify the loan applicants with the highest return on investment and outsources the task to an AI consulting company (i.e., the server). In this scenario, the bank is unable to disclose the raw data of the loan applicants due to privacy and security requirements, whereas the AI consultancy is unwilling to share the implementation of their selection strategy. This scenario can be naturally extended to multiple banks, i.e., multiple banks work collaboratively and aim to identify the most promising loan applicants by utilizing the data of each client privately. In this case, each bank holds its own applicant's data and the overall target is to improve the effectiveness of identifying the most promising loan applicants.
Furthermore, collecting data from devices introduces security threats and vulnerabilities during data transmission and collection, which is a concern for privacy-sensitive applications. In addition, certain raw data may be sensitive and cannot be transferred to third parties due to the General Data Protection Regulation (GDPR). To address these issues, a privacy-preserving framework for data-driven optimization is in high demand, especially when the data is distributed across multiple devices.}}

Privacy-preserving machine learning, including federated learning (FL) \cite{mcmahan2017communication,jin2022book} and learning over encrypted data \cite{cryptoeprint:2014/331} has shown great promise in the past decade. By contrast, only sporadic research has been reported on privacy-preserving data-driven optimization. For example, federated BO \cite{dai2021differentially} is proposed to tackle a global optimization task under a federated framework. To fully utilize the data of clients, Xu \textit{et al.} \cite{xu2021federated_single,xu2021federated_multi} propose a federated acquisition function within a federated data-driven EA framework. Although these studies are inspired by FL, the following challenges arise as the targets of federated learning and federated BO are different. 
\begin{enumerate}
    \item One major challenge in federated optimization is that federated averaging cannot be directly applied to non-parametric surrogate models such as the Gaussian process (GP) widely used in BO, and it is non-trivial to design secure optimization of the acquisition function without having direct access to the predicted objective values of the clients.
    
    \item The second challenge is that the amount of data on each client/user device is often very limited, such as car crash data \cite{jin2009systems} or oil distillation data \cite{han2022surrogate}. This requires to utilize all clients' data, while the privacy of the data must be protected. 
    \item Finally, the secure aggregation schemes in federated learning cannot be directly applied to federated data-driven evolutionary algorithms based on an acquisition function. The reason is that apart from aggregating the weights from all participating clients, the server also needs to propose new query points by optimizing the acquisition function. Thus, secure optimization of the acquisition function without having a direct access to the predicted objective values of each client becomes quintessential. 
\end{enumerate}

Existing work on federated Bayesian optimization and federated data-driven evolutionary algorithm mainly consider the protection of raw data. In \cite{dai2020federated,dai2021differentially}, a surrogate is trained with a set of available hyperparameters and their corresponding validation accuracy that is obtained by the raw data such as images from each client. However, we argue that information such as the hyperparameters and prediction accuracy should not be revealed either. In \cite{xu2021federated_single} and \cite{xu2021federated_multi}, however, the server can directly access the predicted objective values of each client given any set of decision variables.

{\color{black}{We posit that although the surrogate models are unable to accurately predict the real objective values, the predicted objective values obtained based on the model parameters can still reveal private information such as the rank of solutions. This is because the surrogate models reflect the landscape of the surrogates that have been trained using the real function evaluations. To address this issue, we propose to transmit the perturbed predicted objective values of any unobserved solutions instead of directly sharing local model parameters to protect the rank of the newly infilled solutions. Additionally, we argue that the new query points should not be visible to the server, since the server can directly evaluate the quality of all newly infilled solutions, including calculating acquisition function values of all query inputs with the information available from all clients. We propose that the next query input should be suggested on a randomly selected client to prevent the server from knowing the rank of all new solutions and the infilled solutions.}}


{\color{black}{In the context of FL and BO, privacy-preserving mechanisms such as homomorphic encryption (HE) and differential privacy (DP) are widely used for privacy preservation. However, both HE and DP have their own advantages and disadvantages. HE allows for computations to be performed under ciphertext, which provides a high level of security but requires more computational resources and incurs higher communication costs for encryption and decryption. By contrast, DP adds noise to local data or models as a perturbation method, which suffers form a trade-off between accuracy and protection. Given these limitations, this work designs a light-weight privacy-preserving framework that can maintain optimization performance. Our goal is to strike a balance between privacy preservation and optimization performance by leveraging techniques that minimize computational and communication costs while preserving the privacy of sensitive data.}}

Considering that the federated data-driven evolutionary algorithm optimization (FDD-EA) in \cite{xu2021federated_multi} can only protect the raw data, this work proposes a secure privacy-preserving framework based on FDD-EA \cite{xu2021federated_multi} to protect both the raw data and the newly infilled samples. {\color{black}{Our proposed framework incorporates a secure aggregation method based on Diffie-Hellman-based noise, which enables a lightweight design while ensuring strong security. With this approach, our framework can maintain optimization performance without sacrificing privacy.}} The main contributions of this work are:
\begin{itemize}
\item To protect the new infilled solutions, the predicted objective values are masked with Diffie-Hellman-based noise and aggregated on one randomly selected client. Meanwhile, adding different amounts of noise to different individuals in the population changes the relative ranks of the individuals in each client, thereby protecting the predicted values of the clients. 
\item The predicted mean objective value and uncertainty are normalized to reduce the negative influence of the added noise in the estimation of uncertainty when calculating the federated acquisition function value. 

\end{itemize}

The rest of the paper is organized as follows. Section II and Section III present the related work and the preliminaries of this work, respectively. Section IV describes the problem formulation of federated optimization and the security assumptions. The proposed privacy-preserving scheme is detailed in Section V, followed by a security analysis of the proposed algorithm in Section VI and a presentation of the experimental results in Section VII. Finally, Section VIII concludes the paper and proposes future work for the secure federated data-driven evolutionary algorithms. 

\section{Related work}

Here, we briefly discuss the related research, in particular, federated Bayesian optimization and privacy-preserving BO algorithms.



\subsection{Federated Bayesian Optimization}

FL as a new strategy for privacy-preserving machine learning, first proposed by McMahan \textit{et al.} \cite{mcmahan2017communication}, provides a new machine learning paradigm by training local models separately on different clients and aggregating the updated local models on the server. The technology has gained popularity as it can offer a promising solution to data security and privacy protection. Compared with the traditional machine learning techniques, FL can not only improve the learning efficiency but also solve the problem of data silos and protect local data privacy \cite{custers2019eu}. Since the raw data does not leave the owner's local device, FL becomes a popular option for cross-border model training in data-sensitive scenarios, e.g., medical records, personal photo albums, personal voice, and genetic data, just to name a few.
FL can be categorized into horizontal FL, vertical FL, and hybrid FL \cite{yang2019federated}. Horizontal FL shares feature space but has a different sample space in datasets. By contrast, vertical FL shares the same sample space but different feature spaces. Finally, hybrid FL uses a dataset that has a diverse sample and different feature spaces \cite{yang2019federated}. The most typical FL algorithm is FedAvg \cite{mcmahan2017communication}, where the server aggregates the clients' updated models using weighted averaging.

\textcolor{black}{Although it is able to preserve data privacy to a certain degree, the basic FL scheme may be vulnerable to various attacks, such as gradient leakage attacks \cite{zhu2019deep}, inference attacks \cite{rahman2018membership}, and data poisoning attacks \cite{zhang2020online}. An example of data poisoning attack is the pixel-wise backdoor presented in \cite{gu2019badnets}, which results in the canonical example of adversarial and backdoor attacks. To address these issues, researchers have proposed two main categories of privacy-enhancing techniques: encryption methods, such as secure multi-party computation \cite{du2001secure} and homomorphic encryption \cite{rivest1978data}, and data scrambling methods, such as differential privacy \cite{dwork2008differential}. While encryption methods provide strong data security, they often require intensive computational overhead and may be unpractical for edge devices. By contrast, data perturbation methods, which add random noise to the data to prevent attackers from inferring sensitive information about individuals based on the differences in output \cite{shokri2015privacy}, are relatively lightweight. However, they suffer from a trade-off between the model accuracy and privacy preservation.}


{\color{black}
Data-driven evolutionary algorithms \cite{jin2021book} and Bayesian optimization \cite{wang2022survey} can effectively tackle complex and resource-intensive problems, which have attracted increased interest, focusing on constructing effective surrogates and surrogate management strategies in the presence of data paucity. However, most of them are designed in a centralized setting, without considering any privacy issues. For example, in \cite{li2020boosting,li2020data}, the generation of more synthetic data to enrich the original training data requires both the decision variables and the rankings of the top 50\% solutions in the training data. A surrogate-assisted evolutionary algorithm based on multi-population \cite{liu2021surrogate} and two-stage search \cite{zhen2021two} is proposed to strike a balance between exploration and exploitation in handling high-dimensional problems. In both algorithms, a local surrogate model is constructed using a number of top-ranked solutions to enhance the exploitation ability. Luo \textit{et al.} \cite{luo2018surrogate} develop a surrogate-assisted data-driven dynamic optimization algorithm and select the best solutions to construct surrogates to improve the efficiency of finding promising solutions in a rapidly changed environment. In \cite{liu2023surrogate}, a surrogate-assisted two-stage differential evolution is proposed for expensive constrained optimization, in which a repair strategy is introduced to move the infeasible solutions closer to the feasible region and a clustering strategy of feasible solutions is adopted. In the surrogate-assisted cooperative co-evolutionary algorithm, several sub-populations co-evolve to optimize the architecture of liquid state machines \cite{zhou2022}. These algorithms require various types of information in addition to the training data, making them inapplicable to scenarios where the data and the required information are subject to privacy protection.  
} 

There are only a few studies about the federated BO \cite{dai2020federated} and federated data-driven evolutionary optimization \cite{xu2021federated_single}. Federated BO aims to optimize the hyperparameters of a neural network in a federated way by parameterizing the GP model using random Fourier features so that GP models of different clients can be averaged. In addition, Thompson sampling is adopted as the acquisition function with the advantage that its inherent randomness is sufficient to avoid redundant function evaluations when multiple optimization processes are running in parallel, as discussed in \cite{kandasamy2018parallelised}. 
However, the current scheme can only protect the raw data on the clients with the inherent characteristics of FL. It is susceptible to advanced attacks if no additional privacy-preserving techniques are included in FL. To address this issue, the federated BO in \cite{dai2020federated} was extended in \cite{dai2021differentially}, where differential privacy is integrated into federated Thompson sampling. It guarantees user-level privacy by preventing the server from surmising whether a client has been involved in the training or not. 
In addition, the search space is divided into different regions to handle different parameter vectors and improve the optimization efficiency. However, the divisions of the decision variables increase the number of local sub-regions exponentially, making it harder to extend to medium- or high-dimensional problems. Different from federated Thompson sampling with DP, where the optimization of the acquisition function is conducted on the clients, the optimization of the acquisition function is carried out on the server in several other federated BO \cite{sim2021collaborative}. The main benefit of optimizing the acquisition function on the server is that the information in the global and local models can be more easily exploited, and the server usually has sufficient computational resources.

FDD-EA \cite{xu2021federated_single}, {\color{black}as the first work to extend data-driven evolutionary algorithm to a federated setting,} is meant for privacy-preserving single-objective optimization, which replaces the GP with a radial-basis-function network (RBFN) so that federated averaging can be directly adopted. To estimate the uncertainty of the predictions, the standard deviation between the predictions of the local and global model is employed. FDD-EA is extended to multi-objective optimization in \cite{xu2021federated_multi} by obtaining a set of non-dominated solutions using an multi-objective evolutionary algorithm and selecting a subset from these solutions for sampling using k-means.  


The above-mentioned studies do not consider fairness among clients, which may result in some clients being hesitant to involving in training. To achieve both fairness and efficiency among parties, a modified upper confidence bound acquisition function that accounts for fairness is proposed in \cite{sim2021collaborative}. However, this work requires each client to send the real objective values of each query point to the server, which is a huge security risk.

\subsection{Privacy-preserving Mechanisms in Bayesian Optimization}

{\color{black}Canonical BO algorithms suffer from both security and privacy issues.} Adversarial attacks are the most popular attacks, which aim to mislead the BO by making the regret of suggesting appropriate querying points as high as possible. This type of attacks was successfully launched in \cite{han2022adversarial}, which achieved the targeted attacks on GP bandits. {\color{black}Adversial parties can attempt to eavesdrop on private information in an optimization task, such as objective values, decision variables, constraints, or ranking of solutions, depending on what is believed to be sensitive for a special task. Most approaches in privacy-preserving optimization algorithms assume the decision variables and objectives are sensitive. For instance, the rankings of individuals are treated as insensitive in \cite{zhan2021new} and transmitted to a service provider to drive the optimization process with the aim of protecting the real objective values.} Another approach is to directly mask or encrypt the sensitive objective or decision values using DP \cite{dc2008} and homomorphic encryption (HE) \cite{gentry2009fully}, which are two representative additional privacy-preserving techniques to enhance the security level of FL, BO, and evolutionary algorithms. 


DP is considered to be the most secure perturbation-based privacy protection method, which can provide statistical privacy guarantees for individual records and prevent inference attacks on the model without incurring additional computational overhead compared to the encryption methods. However, such systems tend to generate less accurate models due to the added noise in the learning process. In the differential private BO \cite{kusner2015differentially}, Laplace noise is added at the end of the optimization process to ensure the hyperparameters of the neural network are private and near-optimal. However, adding the Laplace noise to the end of the optimization process still has the potential privacy issues of the data leakages. An alternative solution is to add local differential noise to the real objective value in each round of the surrogate update, which can ensure that the best objective value of each client is preserved \cite{nguyen2018privacy}. Recently, Kharkovskii \textit{et al.} proposes to send transformed sets of the decision variables to the service provider, and the BO process is conducted based on the transformed decision variables to protect the query input \cite{kharkovskii2020private}. However, the objective values are still directly exposed to the service provider in \cite{kharkovskii2020private} under this scenario. To protect both decision variables and objective values, Honkela \textit{et al.} \cite{honkela2021gaussian} propose to adopt a sparse GP, which adds Gaussian noise to the corresponding terms of the posterior prediction. By doing so, the decision variables and objective values of the raw data can be kept confidential when conducting inference on unobserved points. Apart from directly adding noise to the objective values like \cite{nguyen2018privacy} in regression tasks, DP is also extended to handle expensive classification problems for data protection. Authors in \cite{xiong2019differential} adopt the Laplace noise generation mechanism and implements it in the expected value of the classification probability, and accomplish classification tasks with an embedded squash function. 

Finally, HE has been introduced in constructing the GP \cite{fenner2020privacy} for protecting data privacy. HE requires that the ciphertext can be directly operated algebraically (typically by addition or multiplication), and the result must be the same as that obtained by using plaintext operations followed by encryption. However, HE cannot be directly applied to federated learning or federated Bayesian optimization. For example, in collaborative scenarios, it remains unclear which client should hold the key. In this regard, an attempt has been made in the literature \cite{RD2019} to propose a distributed key-based HE solution for performing summation computations among untrusted clients. However, the computational cost required to achieve fully HE is extremely high and almost impractical to apply in a federated learning scenario. 
There are only a few pieces of research on how to embed HE into evolutionary optimization or GP modeling for privacy reasons. Earlier work such as \cite{funke2010privacy} implements a HE scheme to achieve a collaborative solving of multi-objective optimization tasks between a producer and a supplier entity by securely encrypting the overall optimization process of a multi-objective evolutionary algorithm. HE is also studied along the line of BO to handle expensive problems, where computational cost is a key consideration. 
To reduce the computational cost, a modular approach to GP modeling is suggested \cite{fenner2020privacy}. As an extension of privacy-preserving GP regression to BO, the GP must be updated iteratively, making it computationally prohibitive. 

To sum up, the aforementioned methods are independent of each other and there are various approaches that can be employed by federated BO to enhance privacy guarantees. Although most existing federated learning systems using DP or HE can achieve good privacy guarantees, some limitations of these methods are currently difficult to overcome. The degradation of model accuracy due to DP methods is more unacceptable for industrial users than the large computational overhead associated with encryption algorithms, as opposed to individual users. Thus, it is necessary to find a novel approach that combines data privacy protection with flexible privacy requirements. 


\section{Preliminaries}

In this section, we will introduce the background of the proposed framework, including the secure aggregation scheme, and the general framework of FDD-EA.

\subsection{Secure Aggregation}

Secure aggregation protocol is a class of secure multi-party computation, where clients submit their updates in such a way that the server will only know the aggregation of those submitted values. It can be implemented, among others, with threshold homomorphic encryption \cite{shi2011privacy} \cite{halevi2011secure},  where the $i$-th client holds its private local dataset $\mathbb{D}_i$, and only when the number of clients reaches a predefined threshold, $\sum{\mathbb{D}_i}$ can be decrypted. One of the most important results of secure aggregation related to private federated training is the protocol of Segal \textit{et al.} \cite{segal2017practical}, which builds on primitives of secret sharing \cite{shamir1979share}, public-key cryptography, and pseudo-random number generation. The main idea of the algorithm is to add random masks to the updates, which will be cancelled out in the aggregation phase. This secure aggregation algorithm consists of two main steps, namely setup for key generation between the clients with Diffie-Hellman key exchange protocol, and secure aggregation for computing the aggregated client data by the server for a given update. In this section, we will introduce the Diffie-Hellman key exchange protocol, which is the building block of secure aggregation.
\begin{figure}
\centering
  \includegraphics[width=0.99\linewidth]{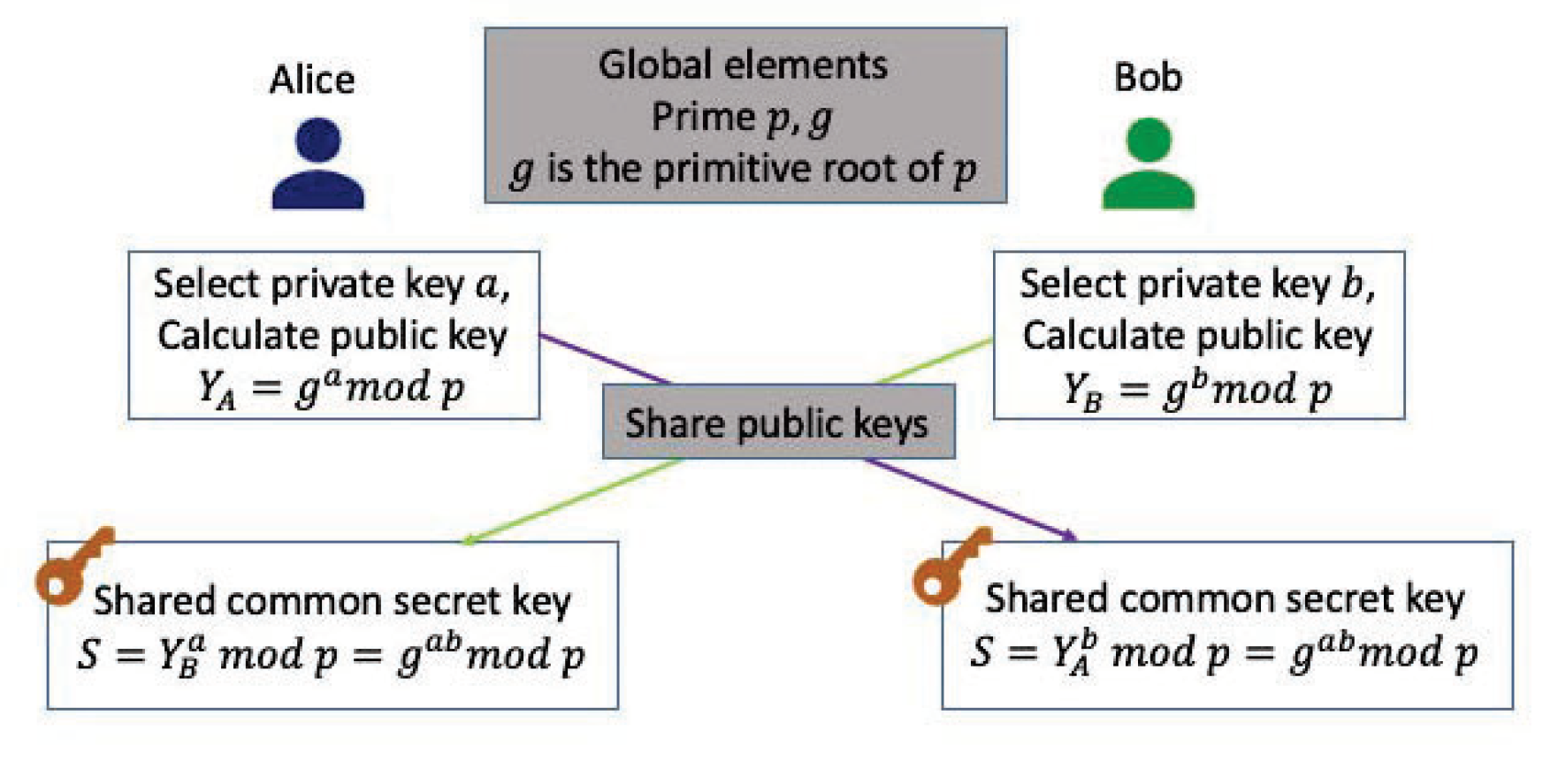}\\
  \caption{Diffie-Hellman Key Exchange Protocol.}
  \label{dh11}
\end{figure} 

In Diffie-Hellman, they use the fact that the multiplicative group $Z^*_p$ for a prime $p$ is \emph{cyclic}. This means that there is an element $g \in Z^*_p$ called \emph{generator}, such that $Z^*_p = \{ 1 ,g ,g^2,g^3,\ldots, g^{p-2} \}$.

\begin{definition}[The Diffie-Hellman (DH) key exchange protocol]
The Diffie-Hellman (DH) key exchange protocol requires a group $(G,\cdot)$ and a generator of the group $g\in G$. 

\begin{enumerate}

\item There are global common elements: random prime number $p$ and a generator $g$.

\item Alice chooses a secret $a$, and calculates public key $Y_A = g^a \pmod p$ and sends it to Bob.

\item Bob chooses a secret $b$, and calculates public key $Y_B = g^b \pmod p$ and send it to Alice.

\item Alice and Bob both compute $S=g^{ab} \pmod{p}$ by computing $Y_B^a$ and $Y_A^a$, respectively.

\item Then they use $S$ as a key to exchange messages using a private key encryption scheme.
\end{enumerate}

\end{definition}

An important question is how to choose $\mathbb{Z}_p$ and the subgroup. The following conditions must be satisfied ($p$ is the order of the group):
\begin{itemize}
    \item $p$ must be large i.e., the size must be 2048 bits at the minimum. If we want to provide a higher security level, it is better to choose a prime with 3072 or 4096 bits.
    \item $p$ must be a safe prime, i.e., $q=(p-1)/2$ is also a prime. Hence the subgroup has an order $q$ that is prime and is large as well. 
\end{itemize}

A random number is a number that cannot be predicted by an observer before it is generated. PRNG refers to an algorithm that processes somewhat unpredictable inputs and generates pseudo-random outputs for cryptographic purposes. It starts from an arbitrary starting state using a random seed (or seed state). After initializing a pseudorandom number generator with the seed state, many random numbers can be generated in a short time and can also be reproduced later.


\subsection{Surrogate Assisted Evolutionary Algorithms}

SAEAs \cite{jin2011surrogate} have been widely used in handling expensive optimization problems. In SAEAs, a set of data containing decision vectors and their objective values is collected before the optimization starts, which is the called raw data. A surrogate model will then be constructed based on the raw data. 
During the search process, new promising solutions are selected according to a model management strategy and queried to update the surrogate \cite{jin2011surrogate} before the search continues. 
Recently, it has become popular in SAEAs to adopt one of the acquisition functions in BO as the model management strategy. In this case, usually the GP model will be used as the surrogate since it can provide a prediction of an unobserved point and a confidence level (uncertainty) of the prediction. Then, like in BO, the acquisition function will be maximized using the EA, and the obtained maximum of the acquisition function is used as the next query point. This procedure continues until a termination condition, usually the allowed maximum number of fitness evaluations, is met. Representative GP-assisted SAEAs for expensive multi-objective optimization include the GP-assisted multi-objective evolutionary algorithm based on decomposition \cite{zhang2009expensive,zhan2017expected}, surrogate-assisted reference vector guided evolutionary algorithm~\cite{chugh2016surrogate}, kriging assisted two archives evolutionary algorithm~\cite{song2021kriging}, and a reference vector guided adaptive model management~\cite{rvmm}. However, they are all designed based on the assumption that both the raw data and the newly queried data are centrally stored on one machine centralized, which cannot directly extended to a federated environment.


\begin{figure}
\centering
  \includegraphics[width=0.8\linewidth]{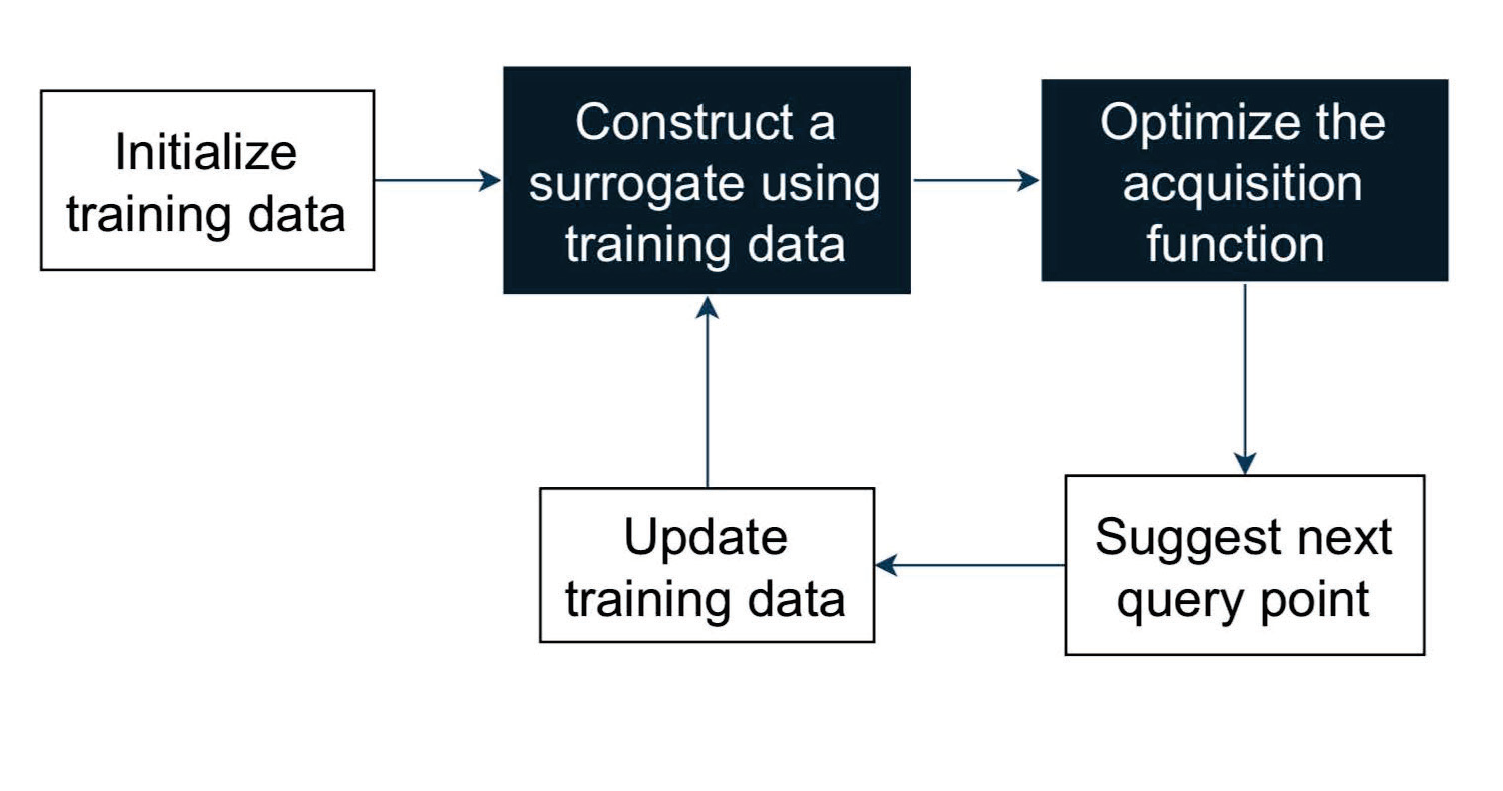}
  \caption{A diagram of a GP-assisted SAEA. The solution set obtained by an SAEA is composed of the initial training data as well as the new query points obtained by optimizing the acquisition function.}
  \label{SAEA}
\end{figure} 

\subsection{A Federated Data-driven Evolutionary Algorithm}

Most data-driven surrogate-assisted EAs are performed on one client, meaning the client can have access to the real objective function, the optimizer, and all training data. However, in many practical scenarios, a number of clients are required to jointly solve the optimization task. Motivated by this, Xu \textit{et al.} \cite{xu2021federated_single,xu2021federated_multi} extended the idea of federated averaging to federated data-driven optimization to deal with single- and multi-objective optimization in a federated environment, which is called federated data-driven evolutionary algorithm, FDD-EA for short. To facilitate the use of federated averaging, the GP in BO is replaced by an RBFN. In this scheme, each client trains a local RBFN and sends the weights of their trained RBFN models to the server. The server then constructs an acquisition function based on the uploaded local models and the aggregated global RBFN to determine the next query point. The federated acquisition function, termed the federated lower confidence bound (FLCB), is constructed by utilizing the predictions of all clients as well as the prediction of the aggregated global RBFN.
\begin{equation}
    \begin{aligned}
    F^{a} = \hat{{\bf{y}}}(x) -t*\sigma(\bf{x})\\
    \hat{{\bf{y}}}({\bf{x}}) = \frac{\sum_{i}^{K} \hat{{\bf{y}}}_i({\bf{x}})/K+\hat{{\bf{y}}}_{s}(\bf{x})}{2}\\
    \sigma({\bf{x}})^2=\frac{1}{K}\left [ \sum_{i}^{K}(\hat{{\bf{y}}}_i({\bf{x}})-\hat{{\bf{y}}}({\bf{x}}) )^2+(\hat{{\bf{y}}}_s({\bf{x}})-\hat{{\bf{y}}}({\bf{x}}) )^2 \right ],    
    \end{aligned}
\label{federated_af}
\end{equation}
\textcolor{black}{where $\hat{{\bf{y}}}({\bf{x}})$ and $\sigma({\bf{x}})$ are the predicted mean objective value and the standard deviation of decision vector $\bf{x}$, respectively. $F^a$ is the federated acquisition function, $\hat{{\bf{y}}}_i({\bf{x}})$ and $\hat{{\bf{y}}}_{s}(\bf{x})$ are the predicted objective values of the $i$-th client and the server ($i \in \left \{1,2,\cdots,K \right \}$), respectively. $K$ is the number of clients and $t$ is a parameter to weight $\hat{{\bf{y}}}({\bf{x}})$ and $\sigma({\bf{x}})$, which is usually a constant.}

\begin{algorithm}
\normalsize
 \SetAlgoLined
 \SetKwData{Left}{left}\SetKwData{This}{this}\SetKwData{Up}{up}
 \SetKwRepeat{doWhile}{do}{while}
 \SetKwFunction{Union}{Union}\SetKwFunction{FindCompress}{FindCompress}
 \SetKwInOut{Input}{Input}\SetKwInOut{Output}{Output}
 \Input{: Number of participating clients $K$, maximum number of generations $t_{m}$,  maximum number of real
objective function evaluations $g_{m}$.}
 \Output{Dataset {$\mathbb{D}_1$, $\mathbb{D}_2$,$\cdots$, $\mathbb{D}_K$}}
 Initialize: Initialize the raw data in client $k$ as $\mathbb{D}_k$ and the number of training data in the initialization stage in all clients is $g_{0}$\; 
 $g$ $\leftarrow$ $g_{0}$\;
 \While{g $<$ $g_{m}$}{

 \tcc{\textbf{On the server side}}
     $\omega_1,\omega_2,\cdots,\omega_N$ $\leftarrow$ Obtain the masked RBFN weights from each client\;
     Initialize $N_{p}$ sets of decision variables as ${\bf{X}}^{'}_{p}$\;
     \For{j=1:$t_{m}$}{
      ${\bf{X}}^{'}_{o}$ $\leftarrow$  Offspring generation using ${\bf{X}}^{'}_{p}$\;
      ${\bf{X}}^{'}$ $\leftarrow$ ${\bf{X}}^{'}_{p} \bigcup {\bf{X}}^{'}_{o}$ \;
      $F^{a}$ $\leftarrow$  Calculate the federated acquisition function of ${\bf{X}}^{'}$ using the RBFN weights of each client. \;
    {\color{black}{  ${\bf{X}}^{'}_{p}$ $\leftarrow$  Environmental selection (${\bf{X}}^{'}, F^{a}$)}} \;
      }
      ${\bf{X}}_q$  $\leftarrow$ Select the next query point from  ${\bf{X}}^{'}_{p}$ \;
      Send ${\bf{X}}_q$ to each client \;
       \tcc{\textbf{On the client side: for each client}}
    $\mathbb{D}_k$ $\leftarrow$ Receive ${\bf{X}}_q$ and add ${\bf{X}}_q$ to the training data $\mathbb{D}_k$ for client $k$\;
     $\omega_k$ $\leftarrow$ Train an RBFN network using $\mathbb{D}_k$ and get the weights for client $k$\;
     Upload the masked weights $\omega_k$ of client $k$ to the server\;
    }
\caption{A general framework of FDD-EA.}
\label{alg1}
\end{algorithm}

The pseudo-code of FDD-EA is given in Algorithm~\ref{alg1}. In FDD-EA, the weights of the locally trained RBFN models are first masked and then sent to the server in each round of surrogate update, as shown in {\color{black}Line 16} in Algorithm~\ref{alg1}. Based on the masked weights of the local RBFNs, the server can obtain a global RBFN by averaging the weights of the local RBFNs. {\color{black}Note that the secure aggregation protocol also ensures that the local model parameters remain hidden from the server. 
} For one decision vector $\bf{x}$, the server can calculate the objective values based on the local RBFNs and the global RBFN, i.e., $\hat{{\bf{y}}}_i({\bf{x}})$ and $\hat{{\bf{y}}}_{s}(\bf{x})$, based on which the value of the FLCB can be calculated for the given decision vector. However, there is a potential privacy risk, because the server can guess the exact objective values of the query point using the transmitted local RBFNs, since the server knows the query points sent to the clients in the previous round. 
As a result, only the raw data can be protected in FDD-EA. By contrast, this work aims to protect both raw data and the new query data points while still being able to using the global surrogate model on the server to perform the optimization.

\begin{figure}
\centering
  \includegraphics[width=0.99\linewidth]{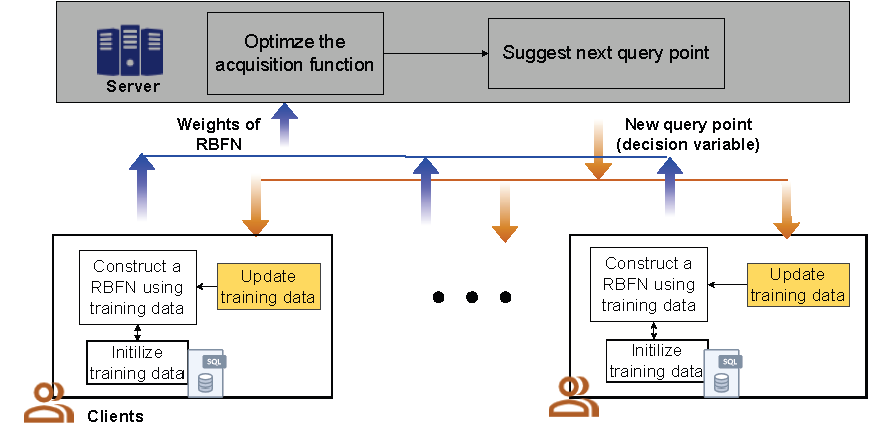}\\
  \caption{A diagram of the federated data-driven evolutionary algorithm. The optimization of the acquisition function is conducted on the server using the predicted objective values of each client, and the prediction of the global surrogate. The global model together with the new query points that are obtained by optimizing the acquisition function will be passed to each client for real objective function evaluations (sampling). On the client side, the received RBFN is further trained based on the local data and the updated weights of the local RBFNs will be transmitted back to the server. }
  \label{FDD}
\end{figure}

\section{Problem Formulation}

Before proposing our framework, we briefly formulate the problem and present the threat models of this work. 

\subsection{Problem Setting}

Our target is to securely handle expensive multi-objective problems on the basis of the FDD-EA framework. We formally formulate an MOP as follows:

\begin{equation}\label{eq1}
\begin{aligned}
\left\{ \begin{array}{l}
\text{min} \; {\bf{f}({\bf{x}})}= (f_{1}({\bf{x}}),f_{2}({\bf{x}}),\cdots , {\color{black}f_{M}}({\bf{x}})),
\\
\text{subject}\; to\; {\bf{x}} = {\color{black}{(x_1,x_2,\dots,x_D)}}, x\in\mathbb{D}_i
\end{array} \right.
\end{aligned}
\end{equation}
where $\mathbf{x}$ is a vector of {\color{black}$D$} decision variables in the decision space $\mathbb{R}^{D}$.  {\color{black}$M$} is the number of objectives and ${\bf{f}}({\bf{x}}) \in \Lambda \subset {\mathbb{R}^{M}}$ is the objective vector with $M$ objectives, $\Lambda $ is the objective space. 

In this work, the objective function $f$ is assumed to be only owned by each client but not the server. Thus, we aim to optimize the objective function $\bf{f}$ in a federated way, and find the global optima of the expensive $\bf{f}$ by utilizing the locally distributed raw data $(\mathbb{D}_1,\mathbb{D}_2,\cdots,\mathbb{D}_K)$ in $K$ clients. Note that each local dataset is changing incrementally with the newly infilled solutions in each round of surrogate update.

In the federated data-driven optimization, a central server is responsible for integrating the information given by each client and recommending new promising solutions to accelerate the optimization of the global function $\bf{f}$. As we mentioned before, our goal is to find the global optima of $\bf{f}$. However, unlike federated learning, it is insufficient in federated BO just to utilize the existing raw data to train an RBFN model on each client for weights sharing. Xu \textit{et al.} \cite{xu2021federated_multi,xu2021federated_single} propose a federated acquisition function on the server to select new query points in order to obtain the global optima as fast as possible. However, their algorithm suffers from the privacy leakage of the new query points and gradients-reconstruction attacks. To combat this huge risk, this work proposes a secure approach to performing optimization of the FLCB acquisition function in FDD-EA.


\subsection{Threat Model and Privacy Guarantees}
{\color{black}{We assume that the user behaviour of every participant of the protocol is semi-honest (sometimes called honest but curious), meaning that the parties have to follow the exact pre-specified protocol, but they may try to learn as much as possible from the information they receive from other parties. Meanwhile, there is no collusion, neither between the server and clients nor between any subset of clients. We also assume that in federated learning, clients are not directly connected to each other, every communication between clients is sent through the server. However, the identities of the clients and central server are verified with digital signature from public-key infrastructure before communication. With public-key infrastructure, it guarantees the identities of the clients and the sent messages by verifying the digital signature. The communication channels between clients and the central server are encrypted over Secure Shell Protocol (SSH) or Hypertext Transfer Protocol (HTTPS). Under this assumption, there is no network or communication hijack, nor the existence of the man-in-the-middle attack. }}

The private information needed to be protected is the local raw dataset $\mathbb{D}_i$ of the clients, and the newly infilled solutions as well.
The original local raw datasets cannot be leaked to other parties or to the public during the cooperated training. The new query points are selected on the client side and are evaluated using the real objective function on the clients.
In the optimization process of the acquisition function, a set of decision variables ${\bf{X}}^{'}$ 
are sent by the server to the clients to obtain the corresponding predicted objective values  $\hat{{\bf{Y}}}= \left \{{\bf{\hat{y}}}^1,{\bf{\hat{y}}}^2,\cdots,{\bf{\hat{y}}}^{N_{p}} \right \}$. The predicted objective values $\hat{{\bf{Y}}}$ of each client and its sequences should be kept confidential to other parties as well.

\textbf{Objective:} We aim to design a privacy-preserving protocol to solve the security problems and protect sensitive information described in this section with the following several security properties: 
\begin{itemize}
    \item \textbf{Correctness.} Given the $i$-th client's local dataset $\mathbb{D}_i$, and the set of decision variables ${\bf{X}}^{'}$, the corresponding objective values $\hat{{\bf{Y}}}$ are error-free in the training. 

    \item \textbf{Confidentiality.} Our framework protects the confidentiality of the data information. Even with malicious clients, and a compromising server, the local data of the clients, their corresponding objective values ${\bf{Y}}^{r}= \left \{{\bf{y}}^{r1},{\bf{y}}^{r2},\cdots,{\bf{y}}^{rN} \right \}$ calculated on the set of decision variables ${\bf{X}}^{r}$, and the ranking sequence of these predicted objective values $\hat{{\bf{Y}}}$ are not disclosed to other parties.

\end{itemize}

\section{Proposed algorithm}
In this section, we introduce the overall framework of the proposed privacy-preserving federated data-driven evolutionary optimization algorithm and the security strategies used in the algorithm.

\subsection{Overall Framework}

\begin{figure}
\centering
  \includegraphics[width=0.99\linewidth]{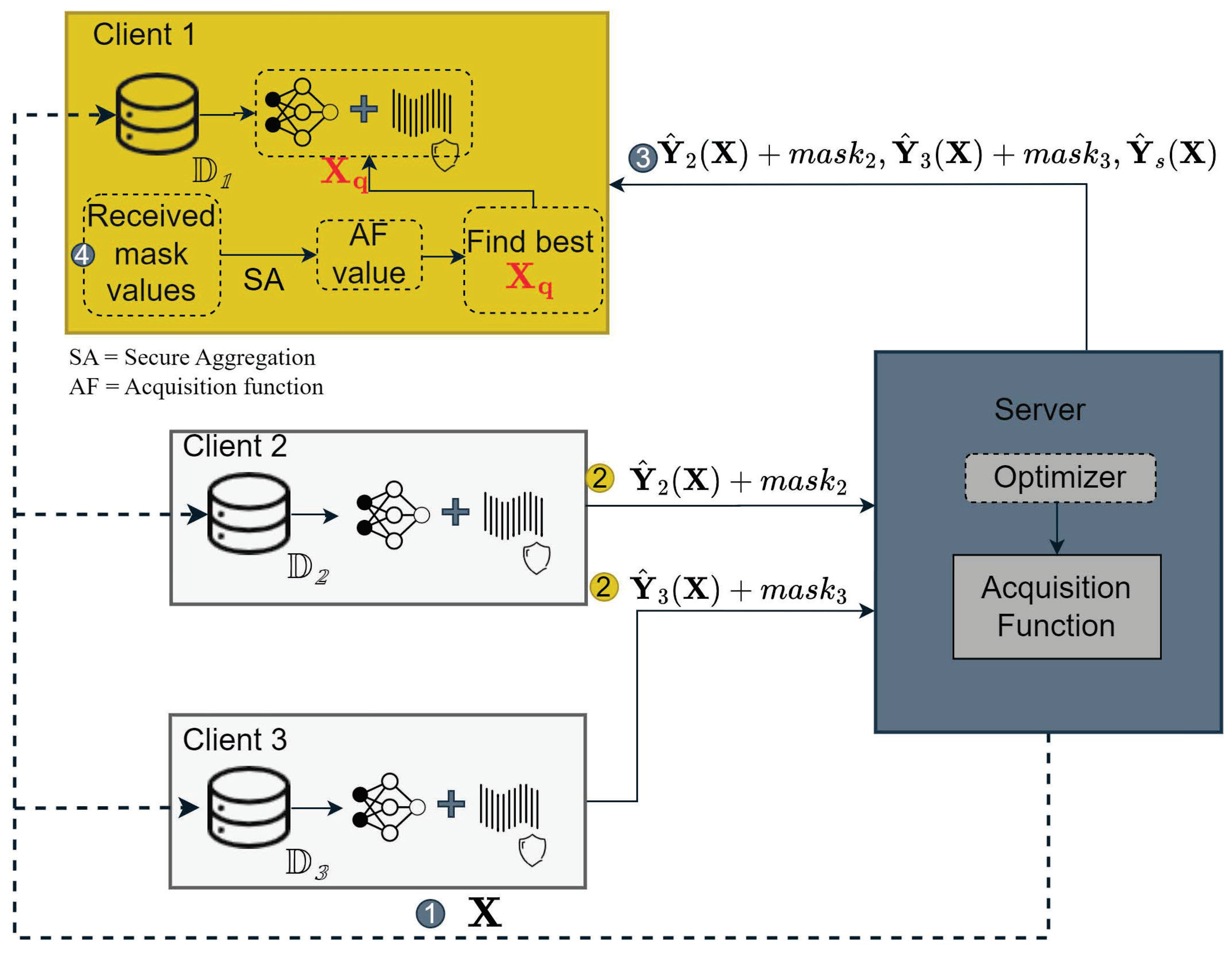}\\
  \caption{An illustrative example of FDD-EA-DH consisting of one server and three clients. \textcircled{1}: In the optimization process of the acquisition function on the server, a set of decision variables $\bf{X}$ are first sent to Clients 2 to 3. \textcircled{2}: After prediction using local models, Clients 2 and 3 send the mask objective values to the server. \textcircled{3}: The server then sends the values from Clients 2 and 3, i.e., $\hat{{\bf{Y}}}_2({\bf{X}})+mask_2,\hat{{\bf{Y}}}_3(X)+mask_3$, as well as the predicted objective values $\hat{{\bf{Y}}}_s(\bf{X})$ using the global model on the server, to Client 1 for secure aggregation. The first three steps will be repeated for $t_{m}$ iterations. \textcircled{4}: At the end of the optimization process, Client 1 selects the next query points ${\bf{X}}_q$ for real objective function evaluations, and after evaluation of ${\bf{X}}_q$ using real function evaluations ${\bf{X}}_q$ is then added to dataset $\mathbb{D}_1$. }
  \label{overview framework}
\end{figure}

In this work, we propose to use a Diffie-Hellman assisted secure-aggregation scheme to protect each client's predicted objective values and ensure that the optimization of the acquisition function is successfully and privately conducted. We call the proposed Diffie-Hellman-assisted FDD-EA, FDD-EA-DH for short. The overall framework is given in Fig. \ref{overview framework} and the pseudo-code of FDD-EA-DH is listed in Algorithm~\ref{alg2} and Algorithm~\ref{alg3}. The architecture includes two parties, the central federated server and multiple federated clients.

According to (\ref{federated_af}), our task is to calculate the federated acquisition value based on the predicted mean values of the local models $\hat{{\bf{y}}}_i({\bf{x}})$ and the predicted mean of the global model $\hat{{\bf{y}}}_s({\bf{x}})$ of each $\bf{x}$. However, due to the security vulnerabilities mentioned above, all the values are exposed to the semi-honest server. For this reason, we do secure aggregation on the client side rather than the server side to design a secure framework, unlike the traditional federated learning setting.

In the optimization process, the central server first sends a set of decision variables $\bf{X}$ generated in the search process to all participating clients, as shown in Step 1 in Fig.~\ref{overview framework}. After receiving the set $\bf{X}$, all clients calculate its predicted objective values based on the local dataset $\mathbb{D}_i$ and send these results $\hat{{\bf{Y}}}_i({\bf{X}})$ back to the server with masked noise. For fairness and secure purposes, a client will be randomly chosen to conduct the calculation of acquisition function values of any unobserved points $\bf{X}$ and the new point will be added into the chosen client database.

Since the estimated objective values are considered to be private, we mask the estimated objective values with Diffie-Hellman-based noise generation before sending to the server, as shown in Step 2 in Fig.~\ref{overview framework}. Since the predicted objective values from the clients are masked, the server will not know the actual predicted objective values of any unobserved points. After that, the server will then send the received masked values to a randomly chosen client (Client 1 in Fig.~\ref{overview framework}), for the calculation of the acquisition function values, as shown in Step 3.

Note that Step 1 to Step 3 will be repeated in each round of surrogate update until the predefined number of iterations $t_{m}$ is exhausted, corresponding to Line 3 to Line 12 in Algorithm~\ref{alg2}. The decision variables ${\bf{X}}^{'}$
are transmitted to all clients and the masked objective values ${\hat{{\bf{Y}}}}^{m}$ will be sent back from the client to the server, ensuring that the actual predicted objective values of each client are not revealed. In the last iteration of one optimization process, i.e., when $j=t_{m}$, the randomly selected client $c_r$ will choose the new query points ${\bf{X}}_q$ based on the FLCB value, corresponding to Lines 8 to 11 in Algorithm~\ref{alg3}. 
The selected query points ${\bf{X}}_q$ will be evaluated using the real objective functions on client $c_r$ to obtain their real objective values. Then, the queried solutions will be added to the training data on client $c_r$. 
Step 4 in Fig.~\ref{overview framework} shows the selection of query points ${\bf{X}}_q$ on Client 1, and after evaluation of ${\bf{X}}_q$ using the real function evaluations, ${\bf{X}}_q$ is added to dataset $\mathbb{D}_1$. Steps 1 to 4 will be repeated until the predefined maximum number of fitness evaluations $g_{m}$ is exhausted.

\subsection{Secure Calculation of Federated Acquisition Function}

In the following, we will present the details of secure aggregation scheme for federated Bayesian optimization, including the mask process of the predicted objective values and the secure calculation of acquisition function values based on the masked objective values. 
\subsubsection{\textbf{Mask the predicted objective values}}
During the optimization process, in order to protect the values of $\hat{{\bf{Y}}}_i({\bf{X}})$ of the $i$-th client from the server and other clients, each client performs a mask operation of $\hat{{\bf{Y}}}_i({\bf{X}})$ by adding some noise. Here we use a variant of Diffie-Hellman key exchange protocol for noise generation. All clients, except for the randomly chosen one, send masked $\hat{{\bf{Y}}}_i({\bf{X}})$ back to the central server. After receiving all the results, the central server will send all the masked objective values $\{\hat{{\bf{Y}}}_i({\bf{X}})+mask_i\}$ and its own predicted value $\hat{{\bf{Y}}}_s({\bf{X}})$ to the randomly chosen client. The corresponding client then conducts a secure aggregation, and finds the best query point ${\bf{X}}_q$ by comparing acquisition function values. This process will be repeated in each round of surrogate update. In the following, we will define how to generate noise and mask $\hat{{\bf{Y}}}_i({\bf{X}})$ with Diffie-Hellman key exchange and how to conduct secure aggregation in the final phase. For each party, it involves the following schemes and processes. 

\textit{Phase 1: Setup and key generations:}
This phase includes the key generation between the clients. Let the set of clients be $C= \left \{1,2, \cdots, K \right \}$. Then the following evaluated parameters will be generated (by the server):

$(G,\cdot),|G|=q, g \in G:G=<g> $ 

(i.e., let $G$ be a cyclic group of $q$ elements with generator $g$. Choose an appropriate group, where the computational Diffie-Hellman problem is hard. Let $i \neq j \in C$ be a fixed pair of clients. Then the following protocol must be run for every pair of clients $i,j$:

\textit{Diffie-Hellman:}
\begin{enumerate}

    \item Client $i$ chooses a random exponent $a_i\in _R \mathbb{Z}_{q-1}$, sends the value $g^{a_i}$ to the server. Then the server sends the pair ($i, g^{a_i}$)  to the client $j$ (the first coordinate is for identification of the sender, it can be omitted if it is solved).
    
    \item Client $j$ chooses a random exponent $a_j\in _R \mathbb{Z}_{q-1}$ and sends the value $g^{a_j}$ to the server. Then the server sends the pair ($j, g^{a_j}$) to the client $i$.
    
    \item The common key is $key_{i,j}=g^{a_ia_j}$.
    
    At the end of this phase, every client $i\in C$ stores a set of common keys:
    
 $K_i=\{{key_{i,1}, key_{i,2}, \cdots , key_{i,i-1}, key_{i,i+1}, \cdots , key_{i,K}}\}$.

\end{enumerate}

\begin{algorithm}
\normalsize
 \SetAlgoLined
 \SetKwData{Left}{left}\SetKwData{This}{this}\SetKwData{Up}{up}
 \SetKwRepeat{doWhile}{do}{while}
 \SetKwFunction{Union}{Union}\SetKwFunction{FindCompress}{FindCompress}
 \SetKwInOut{Input}{Input}\SetKwInOut{Output}{Output}
 \Input{Maximum number of iterations in one optimization process $t_{m}$.}
 \Output{None}
 Initialize a set of decision variables ${\bf{X}}^{'}_{p}$; 
 Generate public parameters $(G,·),|G|=q, g \in G:G=<g> $ (let $G$ be a cyclic group of $q$ elements with generator $g$, and a random salt $r$\; 

     \tcc{In each round of surrogate update}
         The server selects a client $c_r$ randomly \;
     \For{j=1:$t_{m}$}{
      ${\bf{X}}^{'}_{o}$ $\leftarrow$  Offspring generation using ${\bf{X}}^{'}_{p}$\;
      ${\bf{X}}^{'}$ $\leftarrow$ ${\bf{X}}^{'}_{p} \bigcup {\bf{X}}^{'}_{o}$ \;
      Send ${\bf{X}}^{'}$  to each client \;
      $\hat{{\bf{Y}}}^{m}$,$\hat{{\bf{y}}}_s$  $\leftarrow$  Receive the masked objective values $\hat{{\bf{Y}}}^{m}$ corresponding to $X^{'}$ from clients except for client $c_r$ (See Line 14 in Algorithm~\ref{alg3}) {\color{black}and predict objective values $\hat{{\bf{y}}}_s$ of ${\bf{X}}^{'}$ using the global model}\;
      
      Send $\hat{{\bf{Y}}}^{m}$ and {\color{black} the global prediction $\hat{\bf{{\bf{y}}}}_s$} to Client $c_r$\;
      
      ${\bf{F}}^{a}$ $\leftarrow$ Receive FLCB values from Client $c_r$ (See Line 8 in Algorithm~\ref{alg3}) \;
      ${\bf{X}}^{'}_{s}$ $\leftarrow$  Environmental selection (${\bf{X}}^{'},{\bf{F}}^{a}$) \;
      ${\bf{X}}^{'}_{p}$ $\leftarrow$ ${\bf{X}}^{'}_{s}$\;
      }

\caption{Server side of FDD-EA-DH.}
\label{alg2}
\end{algorithm}

\textit{Secure Aggregation}

\begin{enumerate}
    \item The server chooses a random salt $r \in_R{(0,1)}^K$ and sends the same $r$ to every client $i$.
    \item Every client $i \in C$ computes its noise
{\color{black}
\begin{equation}\label{eq3}
    mask_i = \sum_{j <i}P_l(r|key_{i,j})-\sum_{i > j}P_l(r|key_{j,i})
\end{equation} 
    where $P_l$ is a PRNG function, and all clients except for $c_r$ send $\hat{{\bf{Y}}}_i({\bf{x}}) +mask_i$ to the server. $P_l(r|key_{i,j})$ and $P_l(r|key_{j,i}$) is called a noise pair. 
    
    \item The server computes $\hat{\bf{Y}}_m=\sum_{i\ne c_r}(\hat{{\bf{Y}}}_i({\bf{X}}) +mask_i)$ and transmits it to client $c_r$.
    
    \item The client $c_r$ can compute the aggregation of the objective values from
    \begin{equation}\label{eq4}
        \hat{\bf{Y}}_m+\hat{{\bf{Y}}}_{c_r}({\bf{X}}) +mask_{c_r}=\sum_{i\in C}\hat{{\bf{Y}}}_i({\bf{X}}).
        \end{equation}} 
        

\end{enumerate}

\subsubsection{\textbf{Acquisition function value calculation based on the masked objective values}}
 

In FLCB, the uncertainty estimation is based on the predicted objective values of all clients and the server. Since the predicted objective values are masked and influenced by the amounts of noise added, the estimation of uncertainty may become unreliable. To be more specific, the more noise is added, the more likely the uncertainty estimation may become unreliable. On the other hand, however, adding a very small amount of noise may not influence the Pareto dominance relationship between the individuals, leading to possible information leakage to the semi-honest server. Meanwhile, the scale of objective values may change dramatically during the optimization process of FLCB. Thus, we propose to generate relatively a large amount of noise at the beginning of the search process (for minimization) to adapt to the changing objective values. In order to reduce the influence of noise on the calculation of FLCB, we propose to normalize the predicted mean $\hat{{\bf{Y}}}({\bf{X}})$ and uncertainty values ${\bf{\sigma}}^{norm}(\bf{X})$, respectively, before weighting these two terms.
After obtaining the predicted mean value $\hat{{\bf{Y}}}({\bf{X}})$ and uncertainty value $\sigma({\bf{X}})$, we first normalize these two, i.e., $\hat{{\bf{Y}}}^{norm}({\bf{X}})$ and ${\bf{\sigma}}^{norm}(\bf{X})$. 
 , i.e., $\hat{Y}^{norm}({\bf{X}}) = (\hat{{\bf{Y}}}({\bf{X}}) - min(\hat{{\bf{Y}}}({\bf{X}})) / (max(\hat{{\bf{Y}}}({\bf{X}}))- min(\hat{Y}({\bf{X}})))$, ${\bf{\sigma}}^{norm}({\bf{X}}) = ({\bf{\sigma}}({\bf{X}})-min({\bf{\sigma}}({\bf{X}}))/(max({\bf{\sigma}}({\bf{X}}))-min({\bf{\sigma}}({\bf{X}})))$. 
 The FLCB is calculated as follows:

 \begin{equation}
      {\bf{F}}^{a} = \hat{{\bf{Y}}}^{norm}({\bf{X}}) -t*{\bf{\sigma}}^{norm}(\bf{X}).
 \end{equation}
 
\subsection{Individual Ranking Protection}

The inner ranking will not be changed if we add the same amount of noise to the predicted objective values. In this scenario, the semi-honest server can still guess the best query point ${\bf{X}}_q$ from the unchanged ranking. In evolutionary optimization, apart from the exact objective values, the ranking itself can also reveal private information. {\color{black}However, in the proposed algorithm, the server does not have access to any objective value $\hat{\bf{Y}}_i(\bf{X})$ in plain text; it only has access to values that are masked with a noise $mask_i$. It is important to note that each noise $mask_i$ is the sum of $C-1$ pseudo-random strings calculated from Equation \ref{eq3}, which makes it unpredictable to any malicious server in polynomial time (more details can be found in Section \ref{sec_anal}). As a result, the order of the individual rankings is hidden by the noise. Only the order of the masked values is known, which reveals nothing about the original rank.

}

\begin{algorithm}
\normalsize
 \SetAlgoLined
 \SetKwData{Left}{left}\SetKwData{This}{this}\SetKwData{Up}{up}
 \SetKwRepeat{doWhile}{do}{while}
 \SetKwFunction{Union}{Union}\SetKwFunction{FindCompress}{FindCompress}
 \SetKwInOut{Input}{Input}\SetKwInOut{Output}{Output}
 \Input{Number of clients $K$, Number of queried solutions $\mu$.}
 {\color{black}{\Output{The new query points ${\bf{X}}_q$}}}
 Initialize the training data in client $i$ as $\mathbb{D}_i$; 
 Generate secret key $secret key$, calculate its public key $Y_{client} = g^{secret key} \pmod p\ $, generate key pairs $\{K_i={key_{i,1}, key_{i,2}, \cdots , key_{i,i-1}, key_{i,i+1}, \cdots , key_{i,K}}\}$, generate noise based on the key pairs; 
 
     $\omega_i$ $\leftarrow$ Train an RBFN using $\mathbb{D}_i$ and get the weights for each client\;
     \tcc{In each iteration of the optimization process of FLCB}
     ${\bf{X}}^{'}$ $\leftarrow$ Receive the decision variables from the server (See Line 6 in Algorithm~\ref{alg2})\;  
     \textbf{On client $c_r$} \;
     $\hat{{\bf{Y}}}_{c_r}^{m}$ $\leftarrow$ Predict the predicted objective values of ${\bf{X}}^{'}$ and mask them\;
     Receive $\hat{{\bf{Y}}}^{m}$ from the server (See Line 8 in Algorithm~\ref{alg2})\;
     Aggregate the masked objective values of all clients\;
     Calculate the FLCB values of ${\bf{X}}^{'}$ and send to the server\;
    \If{in the last iteration of optimizing FLCB}{
      ${\bf{X}}_q$  $\leftarrow$ Select $\mu$ query points from  ${\bf{X}}^{'}$ \;
      Evaluate ${\bf{X}}_q$ using real objective functions in Client $c_r$ \;
      }
     \textbf{On all clients except client $c_r$} \;
     $\hat{{\bf{Y}}}^{m}$ $\leftarrow$ Predict the predicted objective value of ${\bf{X}}^{'}$, mask it and send to the server\;
\caption{Client side of FDD-EA-DH.}
\label{alg3}
\end{algorithm}

\section{Theoretical Analysis}

In this section, we perform a comprehensive theoretical analysis for FDD-EA-DH algorithm. It includes the aspects of security, communication cost and time complexity. 

\subsection{Security Analysis}\label{sec_anal}

In this subsection, we present a security analysis for our secure aggregation protocol in FDD-EA-DH. {\color{black} The correctness requirement can be rephrased as the correct execution of the optimization algorithm, i.e., the aggregator client is able to compute the aggregation of the objective values. This is exactly the argument in Equation (\ref{eq4}) in the Secure Aggregation protocol, which is a simple consequence of the deterministic property of the PRNG $P_l(.)$ and the fact the pairwise Diffie-Hellman keys are the same, i.e. $key_{i,j}=key_{j,i}$: $$
    \sum_{i\in C}mask_i=\sum_{i\in C}\sum_{j <i}P_l(r|key_{i,j})-\sum_{i > j}P_l(r|key_{j,i})
     =$$ $$=\sum_{i,j\in C}(P_l(r|key_{i,j})-P_l(r|key_{i,j}))=0.$$}

The remaining part of our security framework focuses on the Confidentiality property, which is based on the following technical argument \ref{lem}: 

\begin{lemma}\label{lem}
Let $i,j$ be clients index, $r$ be the random salt, and $P_l$ be the PRNG function. Then the string
    
    $\hat{{\bf{Y}}}_i({\bf{X}}) +mask_i = $
    
    $\hat{{\bf{Y}}}_i({\bf{X}}) + \sum_{j <i}P_l(r|key_{i,j})-\sum_{i > j}P_l(r|key_{j,i})$

is also pseudo-random.
\end{lemma}

As we mentioned in the previous section, we assume our protocol is a secure multiparty computation in the semi-honest setting with all the parties. Meanwhile, there are no collusion attacks among them. Thus, we consider two different scenarios where the server is curious and where the clients are curious. Our protocol ensures that these clients and the central server learn nothing more than their own outputs, and only the chosen aggregator client knows the best query result. The predicted objective values $\hat{{\bf{Y}}}$ corresponding to ${\bf{X}}^{'}$ are confidential to any parties except the client itself. Last but not the least, the added noise changes the sequence of the decision variables, which makes it impossible for the server to guess the best query result from the previously sent ranking. 

\subsubsection{Security in Case of Semi-honest Server}

Consider there is a polynomial adversarial simulator of the semi-honest server, for $Sim_\mathcal{S}$, all $\hat{{\bf{Y}}}_i({\bf{X}}) +mask_i$ are computationally indistinguishable from the real values $\hat{{\bf{Y}}}_i({\bf{x}})$, and it is impossible to get the best query result ${\bf{X}}_q$ based on the shared values $\hat{{\bf{Y}}}_i({\bf{X}}) +mask_i$. 

As we can see from the overall framework of FDD-EA-DH as shown in Fig. \ref{overview framework}, the server has access to all the masked objective values $\{\hat{{\bf{Y}}}_i({\bf{X}})+mask_i\}$ from other clients except for the chosen aggregator client. Since the noise is generated using the PRNG function, it is no longer possible for the simulated adversarial server $Sim_\mathcal{S}$ to guess or reveal the real objective values for the following two reasons. First of all, $Sim_\mathcal{S}$ cannot get access to the masked predicted values of the aggregator client. Consequently, it does not have sufficient information to conduct aggregation. Secondly, the rank of the predicted values of the clients might have been changed. As a result, $Sim_\mathcal{S}$ cannot guess the best query ${\bf{X}}_q$ from the previously sent ranking. In summary, a simulated adversarial server will not be able to get the real objective values $\{\hat{{\bf{Y}}}_i({\bf{X}})\}$ and the best query ${\bf{X}}_q$.

\subsubsection{Security in Case of Semi-honest Aggregator Client}

No private information will be leaked even if the clients are semi-honest, since usually each client only knows its own information. There is an exception, however, when the aggregator client is semi-honest. Assume there is a polynomial adversarial simulator of the semi-honest aggregator client, $Sim_\mathcal{C}$, which can get all the masked predicted values $\hat{{\bf{Y}}}_i({\bf{X}}) +mask_i$ from other clients as well as the noise pairs with other clients, because each pair of clients have a common shared key. However, the aggregator client cannot match the noise pairs based on $\hat{{\bf{Y}}}_i({\bf{X}}) +mask_i$ as the rank sent from the central server is not the same as the previous sent order, and the identities of other paired clients are anonymous to the aggregator client. Thus, the aggregator client cannot exploit any further information from $\hat{{\bf{Y}}}_i({\bf{X}}) +mask_i$ even if it knows the noise pairs. Furthermore, the proposed protocol randomly chooses a client to conduct secure aggregation in each round, significantly reducing the risk of information leakage.


\subsection{Computational and Communication Complexity}
In this subsection, we will conduct a performance analysis of FDD-EA-DH in the aspects of computational and communication complexity. 

\subsubsection{Computational complexity}
Compared with FDD-EA, the main differences in computational cost lie in the key generation at the initialization stage and there is no additional computational cost at each round of surrogate update. Therefore, we focus on analysing the computational cost for key generations, noise generation and secure aggregation in FDD-EA-DH. There is no initialization stage on the server. Thus, we will only analyze the computational cost on clients side. 

\begin{itemize}
    \item On normal clients: We divide the computation cost of each client $i$ into two parts. (1) Performing the $K-1$ key agreements, which takes $O(K-1)$ computation time, (2) Generating noise $P_l(r|key_{i,j})$ with the PRNG function for every other user $j$, which takes $O(K\cdot(K-1))$ time in total. Overall, the computation cost on a normal client is $O(K^2)$.

    \item On the aggregator client: In addition to the communication cost consumed on a normal client, the aggregator client should conduct the computation of aggregation function to remove the masked objective values, which takes $O(K-1)$ time. Thus, the computation cost of the aggregator client is $O(K^2)$ as well.

\end{itemize}

In summary, the total computational cost on a client is $O(K^2)$, no matter whether it is an aggregator client or a normal client.

\subsubsection{Communication complexity}

The communication cost exists when optimizing the acquisition function for key initialization, key generation, and data transmission in FDD-EA-DH. 
Specifically, we consider two measurement metrics for counting the communication complexity in this study, i.e., the number of communication rounds and the size of transmitted data. We assume that prime $p$ has a length more than 64 bits, i.e., every number can be represented with $\log_2p$ bits. In the following, we will analyze the communication complexity on the server side and the client side, respectively.

\begin{itemize}
    \item On the server side: There is no key set up on the server side; thus, we only need to consider the communication cost during the federated optimization process. The server receives masked objective values from $K-1$ clients and forwards all these objective values to a randomly selected client. It is a 2-round communication protocol. Thus, the communication cost on the server side is $2\cdot(K-1) \cdot t_m \cdot N_r \cdot \log_2p$, where $K$ is the number of clients, $t_m$ is the maximum number of iterations for optimizing the acquisition function at each round of surrogate update, and $N_r$ is the number of the rounds of surrogate update.

    \item On normal clients: There are communication costs in both the key generation phase and the data transmission phase. These public parameter exchanges and key generations are set up only once at the beginning of the algorithm. In the key initializing phase, in which two clients are involved during each communication, $K-1$ keys will be generated in total. In Diffie-Hellman key exchange, each client sends the public key to other clients and receives other public keys from other clients. Thus, there are two communication rounds without considering the handshake protocol processes in our case. Thus, the communication cost for key generation of each client is $2 \cdot (K-1)\cdot \log_2p$.  

    During the data transmission phase when optimizing the acquisition function, each client sends the masked objective values back to the server, which only results in one round communication cost of $ t_m \cdot N_r \cdot \log_2p$. Thus, in total the communication cost is $\{2 \cdot (K-1) + t_{m} \cdot N_r\} \cdot \log_2p$.

    \item On the aggregator client: The aggregator client presents a special case. In addition to the communication cost consumed on a normal client, the aggregator client also receives $K-1$ masked objective values from the server. This process requires one communication round only, consuming a communication cost of $(K-1) \cdot t_m \cdot N_r \cdot \log_2p$. Thus, the total communication cost on the aggregator client is $\{2 \cdot (K-1) + ( K-1) \cdot t_m \cdot N_r \}\cdot \log_2p$.
\end{itemize}

In summary, the total communication cost is $(3K-2)\cdot t_m \cdot N_r \cdot \log_2p + 4 \cdot (K-1) \cdot \log_2p$.


\section{Experimental Results}
In this section, we empirically assess the optimization performance of the proposed framework by comparing a number of variants of the federated data-driven evolutionary algorithm with and without privacy protection and security strategies. 
\subsection{Experimental Settings}
We examine the performance of FDD-EA-DH on the DTLZ \cite{deb2005scalable} and WFG \cite{huband2006review} test suite, with the number of objectives $M$ being set to 3, 5, or 10, and the numbers of decision variables $D$ being set to 20. The maximum number of fitness evaluations 
$g_{m}$ for 20-dimensional test instances is set to 219+120 (219 is the number of raw data in each client and 120 is the number of allowed new query points throughout the optimization process), respectively, which is the same as the setting of the number of initially sampled solutions in \cite{rvmm,xu2021federated_multi}, i.e., $11 \cdot D-1$. The number of queried solutions $\mu$ is set to 5. {\color{black}The RBFN model contains one input layer, one hidden layer, and one output layer. The number of local epochs is 20. The learning rate for the clients is 0.06 and the number of the centers of the RBFs is $\sqrt{M+D}+3$. For fairness, the aforementioned parameters are the same as the settings in \cite{xu2021federated_multi}. The number of clients $K$ is set to 4. The population size for the RVEA optimizer is set to 105, 126, and 230 for 3-, 5-, and 10-objective problems, respectively.} The adopted baseline optimizer is the reference vector-assisted evolutionary algorithm \cite{cheng2016reference}.
The inverted generational distance (IGD) \cite{bosman2003balance} is employed as the performance indicator to evaluate the performance of FDD-EA and FDD-EA-DH. {\color{black}The smaller the IGD value is, the better the performance of an algorithm is.}

\begin{table}[htbp]
\renewcommand{\arraystretch}{1.2}
\centering
\caption{The IGD values of solutions obtained by FDD-EA, FDD-EA-DH and two of its variants on the DTLZ test suite. }
\resizebox{0.9\linewidth}{!}{
\begin{tabular}{cccc||cc}
\toprule
Problem                 & M  &FDD-EA    & FDD-EA-DH     & DH-big-wo & DH-big                             \\
\midrule
                        & 3  & 3.50E+02                        & \hl{ 3.45e+2 =} & \hl{ 3.54E+02} & 3.61e+2 =                        \\
                        & 5  & \hl{ 2.57E+02} & 2.62e+2 =                        & \hl{ 2.71E+02} & 2.79e+2 =                        \\
\multirow{-3}{*}{DTLZ1} & 10 & \hl{ 1.47E+02} & 1.50e+2 =                        & 1.58E+02                        & \hl{ 1.54e+2 =} \\
\hline
                        & 3  & 2.75E-01                        & \hl{ 2.24e-1 +} & 7.32E-01                        & \hl{ 2.76e-1 +} \\
                        & 5  & \hl{ 4.03E-01} & 6.52e-1 -                        & 8.25E-01                        & \hl{ 6.37e-1 +} \\
\multirow{-3}{*}{DTLZ2} & 10 & \hl{ 7.60E-01} & 1.04e+0 -                        & \hl{ 1.04E+00} & 1.04e+0 =                        \\
\hline
                        & 3  & \hl{ 9.98E+02} & 1.07e+3 -                        & 1.06E+03                        & \hl{ 1.04e+3 =} \\
                        & 5  & 8.83E+02                        & \hl{ 8.72e+2 =} & 9.29E+02                        & \hl{ 9.23e+2 =} \\
\multirow{-3}{*}{DTLZ3} & 10 & \hl{ 4.89E+02} & 4.96e+2 =                        & \hl{ 5.05E+02} & 5.15e+2 =                        \\
\hline
                        & 3  & 1.00E+00                        & \hl{ 9.71e-1 =} & 1.17E+00                        & \hl{ 9.92e-1 +} \\
                        & 5  & \hl{ 1.09E+00} & 1.12e+0 =                        & 1.30E+00                        & \hl{ 1.13e+0 +} \\
\multirow{-3}{*}{DTLZ4} & 10 & \hl{ 1.12E+00} & 1.15e+0 -                        & 1.19E+00                        & \hl{ 1.17e+0 =} \\
\hline
                        & 3  & 2.28E-01                        & \hl{ 1.68e-1 +} & 6.45E-01                        & \hl{ 2.34e-1 +} \\
                        & 5  & \hl{ 2.15E-01} & 2.29e-1 =                        & 5.73E-01                        & \hl{ 2.27e-1 +} \\
\multirow{-3}{*}{DTLZ5} & 10 & \hl{ 1.76E-01} & 3.79e-1 -                        & 4.21E-01                        & \hl{ 3.71e-1 +} \\
\hline
                        & 3  & \hl{ 1.54E+01} & 1.54e+1 =                        & \hl{ 1.54E+01} & 1.54e+1 =                        \\
                        & 5  & \hl{ 1.37E+01} & 1.37e+1 =                        & \hl{ 1.36E+01} & 1.36e+1 =                        \\
\multirow{-3}{*}{DTLZ6} & 10 & 9.30E+00                        & \hl{ 9.30e+0 =} & 9.24E+00                        & \hl{ 9.21e+0 =} \\
\hline
                        & 3  & \hl{ 7.83E+00} & 7.86e+0 =                        & 8.12E+00                        & \hl{ 7.91e+0 =} \\
                        & 5  & 1.37E+01                        & \hl{ 1.31e+1 =} & \hl{ 1.29E+01} & 1.33e+1 =                        \\
\multirow{-3}{*}{DTLZ7} & 10 & \hl{ 2.25E+01} & 2.53e+1 -                        & 2.43E+01                        & \hl{ 2.38e+1 =} \\
\hline
\multicolumn{2}{c}{+/-/=}    &                                 & 2/6/13                           &                                 & 7/0/14            \\              
\bottomrule
\end{tabular}}
\label{dh}
\end{table}

\subsection{Performance Comparison with/out Diffie-Hellman}

Table~\ref{dh} lists the IGD values of solutions obtained by FDD-EA and FDD-EA-DH on the 20-dimensional DTLZ test instances. As shown in the table, FDD-EA-DH outperforms FDD-EA on 2 out of 21 test instances and is outperformed by FDD-EA on 6 test instances. We also conduct comparative experiments on the WFG test instances, and the results are shown in Table~\ref{dh_wfg}. It is interesting to find that FDD-EA-DH is able to obtain slightly better performance than FDD-EA. Specifically, FDD-EA-DH wins on 9 out of 27 test instances but is outperformed by FDD-EA on 7 out of 27 test instances.
Overall, FDD-EA-DH achieves similar performance to FDD-EA on the majority of the test instances, indicating that FDD-EA-DH is able to provide a secure way of conducting federated data-driven optimization with only a little sacrifice on performance. Note that both the server and the clients are assumed to be semi-honest in this work.

\begin{table}[htbp]
\renewcommand{\arraystretch}{1.2}
\centering
\caption{The IGD values of solutions obtained by FDD-EA, FDD-EA-DH and two of its variants on the WFG test suite. }
\resizebox{0.9\linewidth}{!}{
\begin{tabular}{cccc||cc}
\toprule
Problem                 & M  &FDD-EA    & FDD-EA-DH     & DH-big-wo & DH-big                             \\
\midrule

                       & 3  & \hl{ 2.47E+00} & 2.48e+0 =                        & \hl{ 2.46E+00} & 2.47e+0 =                        \\
                       & 5  & \hl{ 2.65E+00} & 2.67e+0 =                        & \hl{ 2.65E+00} & 2.67e+0 =                        \\
\multirow{-3}{*}{WFG1} & 10 & \hl{ 3.44E+00} & 3.44e+0 =                        & \hl{ 3.44E+00} & 3.44e+0 =                        \\
\hline
                       & 3  & 1.05E+00                        & \hl{ 9.84e-1 +} & 1.05E+00                        & \hl{ 9.78e-1 +} \\
                       & 5  & 1.52E+00                        & \hl{ 1.46e+0 =} & 1.53E+00                        & \hl{ 1.52e+0 =} \\
\multirow{-3}{*}{WFG2} & 10 & 3.72E+00                        & \hl{ 3.63e+0 =} & 3.81E+00                        & \hl{ 3.57e+0 =} \\
\hline
                       & 3  & 6.38E-01                        & \hl{ 4.85e-1 +} & 6.90E-01                        & \hl{ 5.10e-1 +} \\
                       & 5  & 8.49E-01                        & \hl{ 5.97e-1 +} & 9.53E-01                        & \hl{ 6.28e-1 +} \\
\multirow{-3}{*}{WFG3} & 10 & 1.37E+00                        & \hl{ 8.76e-1 +} & 1.50E+00                        & \hl{ 8.69e-1 +} \\
\hline
                       & 3  & 1.00E+00                        & \hl{ 9.49e-1 =} & 9.94E-01                        & \hl{ 9.80e-1 =} \\
                       & 5  & 2.27E+00                        & \hl{ 2.20e+0 +} & 2.29E+00                        & \hl{ 2.28e+0 =} \\
\multirow{-3}{*}{WFG4} & 10 & 9.49E+00                        & \hl{ 9.37e+0 =} & 9.59E+00                        & \hl{ 9.55e+0 =} \\
\hline
                       & 3  & 9.03E-01                        & \hl{ 8.79e-1 +} & 8.96E-01                        & \hl{ 8.84e-1 =} \\
                       & 5  & \hl{ 1.69E+00} & 1.71e+0 =                        & 1.68E+00                        & \hl{ 1.64e+0 =} \\
\multirow{-3}{*}{WFG5} & 10 & \hl{ 6.67E+00} & 6.94e+0 -                        & 7.08E+00                        & \hl{ 7.07e+0 =} \\
\hline
                       & 3  & 9.81E-01                        & \hl{ 9.52e-1 +} & 1.04E+00                        & \hl{ 9.60e-1 +} \\
                       & 5  & 1.90E+00                        & \hl{ 1.82e+0 +} & 1.92E+00                        & \hl{ 1.88e+0 =} \\
\multirow{-3}{*}{WFG6} & 10 & \hl{ 7.39E+00} & 7.64e+0 -                        & \hl{ 7.33E+00} & 7.38e+0 =                        \\
\hline
                       & 3  & \hl{ 8.13E-01} & 8.59e-1 -                        & \hl{ 8.73E-01} & 8.78e-1 =                        \\
                       & 5  & \hl{ 1.81E+00} & 1.89e+0 -                        & \hl{ 1.83E+00} & 1.93e+0 -                        \\
\multirow{-3}{*}{WFG7} & 10 & \hl{ 7.84E+00} & 8.07e+0 -                        & \hl{ 8.06E+00} & 8.24e+0 =                        \\
\hline
                       & 3  & 1.11E+00                        & \hl{ 1.04e+0 +} & 1.08E+00                        & \hl{ 1.04e+0 +} \\
                       & 5  & 2.01E+00                        & \hl{ 1.99e+0 =} & 2.01E+00                        & \hl{ 1.99e+0 =} \\
\multirow{-3}{*}{WFG8} & 10 & 7.77E+00                        & \hl{ 7.74e+0 =} & \hl{ 7.67E+00} & 7.74e+0 =                        \\
\hline
                       & 3  & \hl{ 8.01E-01} & 8.21e-1 =                        & 9.61E-01                        & \hl{ 8.81e-1 +} \\
                       & 5  & \hl{ 1.75E+00} & 2.07e+0 -                        & \hl{ 2.09E+00} & 2.12e+0 =                        \\
\multirow{-3}{*}{WFG9} & 10 & \hl{ 7.33E+00} & 8.07e+0 -                        & \hl{ 7.81E+00} & 8.27e+0 -                        \\
\hline
\multicolumn{2}{c}{+/-/=}   &                                 & 9/7/11                           &                                 & 7/2/18      \\
\bottomrule
\end{tabular}}
\label{dh_wfg}
\end{table}

{\color{black}
\subsection{Performance Comparison with a Privacy-preserving Bayesian Optimization Algorithm}

In this subsection, we aim to extend a DP-based privacy-preserving Bayesian optimization (BO) algorithm proposed by Nguyen \textit{et al.} \cite{nguyen2018privacy} to the federated multi-objective setting. We termed this variant as EPPBO-KRVEA and compared its performance with FDD-EA-DH. The main concept behind EPPBO is to introduce carefully designed noise to the actual objective values of newly filled solutions to attain the differential privacy of the current best solution. However, EPPBO is only designed to handle single-objective optimization problems. Therefore, we integrated the noise generation approach of EPPBO into the KRVEA \cite{chugh2016surrogate} framework by adding differential privacy noise to each objective value of the newly filled solutions. The results of our experiments on DTLZ test instances are presented in Table \ref{EPPBO_KRVEA}. The results show that our approach outperforms EPPBO-KRVEA on 11 out of 21 test instances and is outperformed by EPPBO-KRVEA on only 3 out of 21 test instances, indicating that our approach is competitive against state-of-the-art DP-based privacy-preserving BO algorithms.

\begin{table}[htbp]
\renewcommand{\arraystretch}{1.2}
\centering
\caption{{\color{black}The mean IGD value and the standard deviation values of the solutions obtained by FDD-EA-DH and EPPBO-KRVEA on the DTLZ test suite over 20 independent runs.}}
\resizebox{0.75\linewidth}{!}{
\begin{tabular}{cccc}
\toprule
Problem&$M$&FDD-EA-DH&EPPBO\_KRVEA\\
\midrule
\multirow{3}{*}{DTLZ1}&3&\hl{3.45e+2 (2.49e+1)}&3.69e+2 (2.25e+1) $-$\\
&5&\hl{2.62e+2 (2.47e+1)}&2.78e+2 (3.48e+1) $\approx$\\
&10&1.50e+2 (2.03e+1)&\hl{1.37e+2 (3.84e+1) $\approx$}\\
\hline
\multirow{3}{*}{DTLZ2}&3&\hl{3.16e-1 (3.05e-2)}&7.59e-1 (1.11e-1) $-$\\
&5&\hl{7.08e-1 (6.62e-2)}&8.30e-1 (8.35e-2) $-$\\
&10&\hl{1.03e+0 (3.39e-2)}&1.03e+0 (8.10e-2) $\approx$\\
\hline
\multirow{3}{*}{DTLZ3}&3&1.07e+3 (1.18e+2)&\hl{1.03e+3 (1.14e+2) $\approx$}\\
&5&8.72e+2 (9.81e+1)&\hl{8.55e+2 (7.20e+1) $\approx$}\\
&10&\hl{4.96e+2 (6.70e+1)}&5.51e+2 (8.32e+1) $-$\\
\hline
\multirow{3}{*}{DTLZ4}&3&\hl{9.71e-1 (6.61e-2)}&1.27e+0 (1.30e-1) $-$\\
&5&\hl{1.12e+0 (3.47e-2)}&1.33e+0 (8.30e-2) $-$\\
&10&\hl{1.15e+0 (4.72e-2)}&1.19e+0 (4.88e-2) $-$\\
\hline
\multirow{3}{*}{DTLZ5}&3&\hl{2.15e-1 (2.74e-2)}&7.06e-1 (1.26e-1) $-$\\
&5&\hl{2.29e-1 (2.42e-2)}&6.82e-1 (6.91e-2) $-$\\
&10&\hl{3.79e-1 (4.24e-2)}&4.13e-1 (4.54e-2) $-$\\
\hline
\multirow{3}{*}{DTLZ6}&3&1.54e+1 (1.82e-1)&\hl{1.34e+1 (1.10e+0) $+$}\\
&5&1.37e+1 (1.35e-1)&\hl{1.26e+1 (7.84e-1) $+$}\\
&10&9.30e+0 (7.92e-2)&\hl{8.98e+0 (4.48e-1) $+$}\\
\hline
\multirow{3}{*}{DTLZ7}&3&7.86e+0 (5.46e-1)&\hl{6.61e+0 (2.61e+0) $\approx$}\\
&5&1.31e+1 (1.26e+0)&\hl{1.20e+1 (2.79e+0) $\approx$}\\
&10&\hl{2.27e+1 (2.25e+0)}&2.57e+1 (3.38e+0) $-$\\
\hline
\multicolumn{3}{c}{$+/-/\approx$}&3/11/7\\
\bottomrule
\end{tabular}}
\label{EPPBO_KRVEA}
\end{table}

}
\subsection{Influence of Normalization on FLCB}
 When the added amount of noise to one solution is very big relative to the solution's predicted objective values, the estimated FLCB values may be heavily influenced. To address this issue, this work proposes to normalize the predicted mean and uncertainty values in calculating FLCB. Here, we further compare the performance of FDD-EA-DH with and without normalization when the added amount of noise is extremely big. The two variants of FDD-EA-DH with and without normalization when very big noise is added are denoted by DH-big and DH-big-wo. In Table~\ref{dh} and Table~\ref{dh_wfg}, we find that DH-big can always outperform DH-big-wo on both DTLZ and WFG test instances, indicating that the normalization strategy has effectively enhanced the performance of FDD-EA-DH when the added amount of noise is very large relative to the solutions' objective values. Thus, we conclude that FDD-EA-DH can well scale to problems with different objective scales.

{\color{black}
\subsection{Performance Comparison with Homomorphic Encryption based FDD-EA}
To showcase the efficiency of FDD-EA-DH, we conduct an implementation of the FLCB acquisition function using additive homomorphic encryption (HE) within the FDD-EA framework, resulting in a variant called FDD-EA-HE. In our particular experiment, we use the Paillier \cite{paillier1999public} partially HE algorithm. Each client encrypts the predicted value with the server's public keys and sends the resulting ciphertext to the server. The server then aggregates the ciphertext received from all clients and decrypts it using its private keys. 

Table \ref{he} demonstrates that FDD-EA-HE outperforms FDD-EA-DH on 13 out of 48 test instances while underperforms on 5 out of 48 test instances. These results suggest that FDD-EA-DH performs only slightly worse than FDD-EA-HE. Although FDD-EA-HE does not add noise when calculating the standard deviations for the optimization problems, which may slightly improve its performance, it is important to note that FDD-EA-HE incurs a significantly higher computational cost compared to FDD-EA-DH. This is because FDD-EA-HE employs homomorphic encryption, which is computationally intensive as it operates under ciphertext. 

As explained in Section VI.B, the computational and communication complexity of FDD-EA-DH is mainly determined by the key exchange. The computational cost of FDD-EA-DH is proportional to the square of the number of clients ($K$), with a complexity of $O(K^2)$. On the other hand, FDD-EA-HE encrypts the plaintext, and the computational cost is mainly determined by the encryption and decryption of the message size. For each client, it encrypts the predicted value and sends the resulting ciphertext to the server. Therefore, the computational complexity for each client is $O(M^2)$, where $M$ is the size of the message. It is worth noting that the size of the message is much larger than the number of clients.

\begin{table}[htbp]
\renewcommand{\arraystretch}{1.2}
\centering
\caption{{\color{black}The IGD values of the solutions obtained by FDD-EA-DH and FDD-EA-HE on the DTLZ and WFG test suites.} }
\resizebox{0.9\linewidth}{!}{
\begin{tabular}{cccc||cccc}
\toprule
Problem                 & M                    & FDD-EA-DH                                     & FDD-EA-HE                                & Problem                & M  & FDD-EA-DH                                    & FDD-EA-HE                                 \\
\midrule
                        & 3                    & \hl{ 3.45e+2 } & 3.55e+2  =                        &                        & 3  & 9.84e-1                         & \hl{ 9.63e-1  =} \\
                        & 5                    & 2.62e+2                        & 2.62e+2  =                        &                        & 5  & 1.46e+0                         & \hl{ 1.43e+0  =} \\
                        
\multirow{-3}{*}{DTLZ1} & 10                   & 1.50e+2                         & 1.53e+2  =                        & \multirow{-3}{*}{WFG2} & 10 & \hl{ 3.63e+0 } & 3.66e+0  =                        \\
\hline
                        & 3                    & 3.16e-1                         & \hl{ 3.14e-1  =} &                        & 3  & 4.85e-1                        & \hl{ 4.73e-1  =} \\
                        & 5                    & 7.08e-1                        & \hl{ 5.60e-1  +} &                        & 5  & 5.97e-1                         & \hl{ 5.67e-1  +} \\
\multirow{-3}{*}{DTLZ2} & 10                   & 1.03e+0                         & \hl{ 8.74e-1  +} & \multirow{-3}{*}{WFG3} & 10 & \hl{ 8.76e-1 } & 9.76e-1  -                        \\
\hline
                        & 3                    & 1.07e+3                         & \hl{ 1.05e+3  =} &                        & 3  & \hl{ 9.49e-1 } & 9.85e-1  =                        \\
                        & 5                    & 8.72e+2                         & \hl{ 8.68e+2  =} &                        & 5  & \hl{ 2.20e+0 } & 2.31e+0  -                        \\
\multirow{-3}{*}{DTLZ3} & 10                   & 4.96e+2                         & 5.04e+2  =                        & \multirow{-3}{*}{WFG4} & 10 & \hl{ 9.37e+0 } & 9.69e+0  -                        \\
\hline
                        & 3                    & 9.71e-1                         & \hl{ 9.55e-1  +} &                        & 3  & \hl{ 8.79e-1 } & 9.19e-1 -                        \\
                        & 5                    & 1.12e+0                         & \hl{ 1.11e+0  =} &                        & 5  & 1.71e+0                         & 1.70e+0 =                        \\
\multirow{-3}{*}{DTLZ4} & 10                   & 1.15e+0                         & \hl{ 1.14e+0  =} & \multirow{-3}{*}{WFG5} & 10 & 6.94e+0                         & 6.73e+0  =                        \\
\hline
                        & 3                    & 2.15e-1                        & \hl{ 1.81e-1  +} &                        & 3  & 9.52e-1                         & \hl{ 8.93e-1  +} \\
                        & 5                    & 2.29e-1                        & \hl{ 2.03e-1  +} &                        & 5  & \hl{ 1.82e+0 } & 1.83e+0=                        \\
\multirow{-3}{*}{DTLZ5} & 10                   & 3.79e-1                         & \hl{ 1.60e-1  +} & \multirow{-3}{*}{WFG6} & 10 & 7.64e+0                         & 7.41e+0  +                        \\
\hline
                        & 3                    & 1.54e+1                         & 1.53e+1  =                        &                        & 3  & 8.59e-1                        & \hl{ 7.94e-1  +} \\
                        & 5                    & 1.37e+1                        & 1.37e+1  =                        &                        & 5  & 1.89e+0                        & 1.85e+0  =                        \\
\multirow{-3}{*}{DTLZ6} & 10                   & 9.30e+0                       & \hl{ 9.26e+0  =} & \multirow{-3}{*}{WFG7} & 10 & 8.07e+0                         & 8.16e+0  =                        \\
\hline
                        & 3                    & 7.86e+0                       & \hl{ 7.85e+0  =} &                        & 3  & \hl{ 1.04e+0 } & 1.06e+0  =                        \\
                        & 5                    & 1.31e+1                       & \hl{ 1.25e+1  =} &                        & 5  & \hl{ 1.99e+0 } & 2.00e+0  =                        \\
\multirow{-3}{*}{DTLZ7} & 10                   & \hl{ 2.27e+1 } & 2.42e+1  -                        & \multirow{-3}{*}{WFG8} & 10 & \hl{ 7.74e+0 } & 7.75e+0  =                        \\
\hline
                        & 3                    & 2.48e+0                        & 2.47e+0  =                        &                        & 3  & 8.21e-1                        & \hl{ 7.83e-1  +} \\
                        & 5                    & 2.67e+0                        & \hl{ 2.66e+0  =} &                        & 5  & 2.07e+0                         & 1.89e+0  +                        \\
\multirow{-3}{*}{WFG1}  & 10                   & 3.44e+0                      & \hl{ 3.43e+0  =} & \multirow{-3}{*}{WFG9} & 10 & 8.07e+0                       & \hl{ 7.11e+0  +} \\
\hline
\multicolumn{1}{l}{}    & \multicolumn{1}{l}{} & \multicolumn{1}{l}{}                     & \multicolumn{1}{l}{}                       & \multicolumn{2}{c}{+/-/=}   &                                          & 13/5/30  \\
\bottomrule
\end{tabular}}
\label{he}
\end{table}

}
\subsection{Analysis of Individual Rank Protection}


{\color{black} By adopting the masked objective values in our work, we can assure that the masked objective values are significantly different from the approximated objective values so that both the predicted objective values in a population and the rank of all individuals in a population are significantly changed. To verify this, we use the rank correlation proposed in \cite{jin2011surrogate} to measure the monotonic relationship between the rank of the approximated objective values and that of the masked objective values. The ranking correlation is defined as follows.
 \begin{equation}
      \rho^{rank} = 1 - \frac{6\sum_{l=0}^{\lambda}d_l^2}{\lambda(\lambda^2-1)},
 \end{equation}
where $\lambda$ is the number of individuals in one population and $d_l$ is the differences between the ranks of $l$-individual based on the predicted objective values and the masked objective values. The range of $\rho^{rank}$ is within [-1,1]. The higher the value of $\rho^{rank}$, the stronger the monotonic relationship with a positive slope between the ranks of the two variables. 

In this subsection, we calculate the mean values of $\rho^{rank}$ of each objective of the solutions over the whole optimization process on 3-objective DTLZ2, DTLZ5, WFG1 and WFG5 test instances. The rank correlations on these test instances are 0.0067, 0.0039, -0.0014, and 0.0019, respectively. We can see that the rank correlations are nearly zero, indicating that the ranks of the masked objective values have become significantly different from the ranks of the approximated objective values on these test problems. Thus, we conclude that our strategy can protect not only the predicted objective values of the solutions of each client but also their relative ranks. }

\section{Conclusion and future work}
In this paper, we propose a privacy-preserving federated data-driven optimization framework based on the Diffie-Hellman-assisted secure aggregation, called FDD-EA-DH, to securely handle data-driven expensive multi-objective problems. FDD-EA-DH provides a secure way of optimizing the federated acquisition function without revealing the users' predicted objective values during the search process. Different from the existing work, not only the raw data, but also the newly infilled solutions are protected. By adding different amounts of noise to different local predictions, the ranking of the predictions of the clients can also be protected, which is of great importance since an attacker usually cares more about the relative rankings between the clients rather than their exact objective values in the context of optimization. Our experiments show that the implemented security strategy leads to only negligible performance degradation in federated optimization. 

Many challenges remain to be resolved in the proposed secure federated data-driven optimization. For example, instead of randomly selecting a client in each round, there might be a better way to ensure that all participating clients can simultaneously obtain the new query points suggested by the server. Also, the distribution of the raw data on each client may be strongly non-iid, calling for a more sophisticated model aggregation method. {\color{black}{Furthermore, we intend to broaden our security setting to incorporate malicious clients or compromised servers, which is a more realistic scenario that necessitates additional security measures on top of the proposed framework.}} Finally, computationally more efficient encryption-based protocols should also be studied to strike a good balance between performance and security.


\bibliography{main}

\begin{IEEEbiography}
    [{\includegraphics[width=1in,height=1.25in,clip,keepaspectratio]{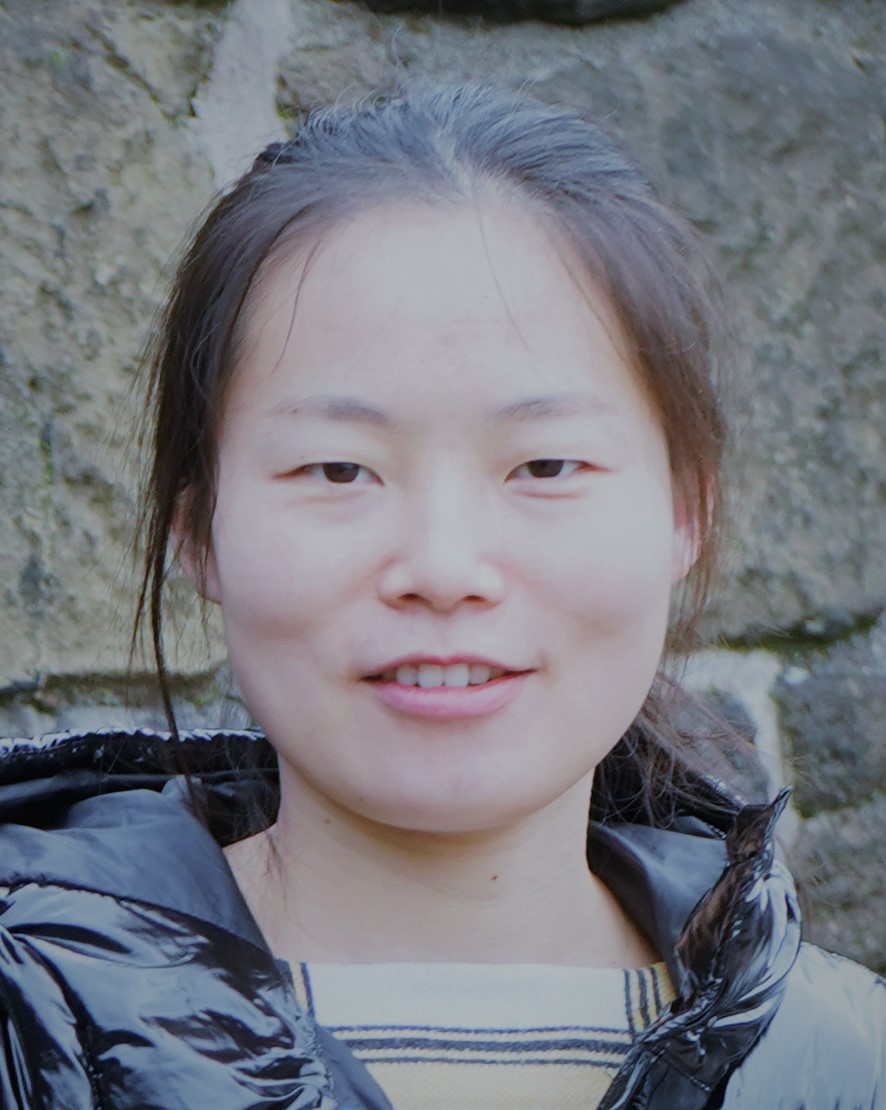}}]{Qiqi Liu} received the B.Eng. degree in industrial engineering from Xi'an University of Science and Technology and the M.E. degree in information and communication engineering from Shenzhen University in 2013 and 2016, respectively. She received her Ph.D. degree in computer science at University of Surrey in 2022. Her current research interests include evolutionary many-objective optimization, surrogate-assisted evolutionary optimization, federated Bayesian optimization, and federated learning. She is a regular reviewer of IEEE Transactions on Evolutionary Computation, Complex \& Intelligent Systems, and Swarm and Evolutionary Computation.
\end{IEEEbiography}

\begin{IEEEbiography}
    [{\includegraphics[width=1in,height=1.25in,clip,keepaspectratio]{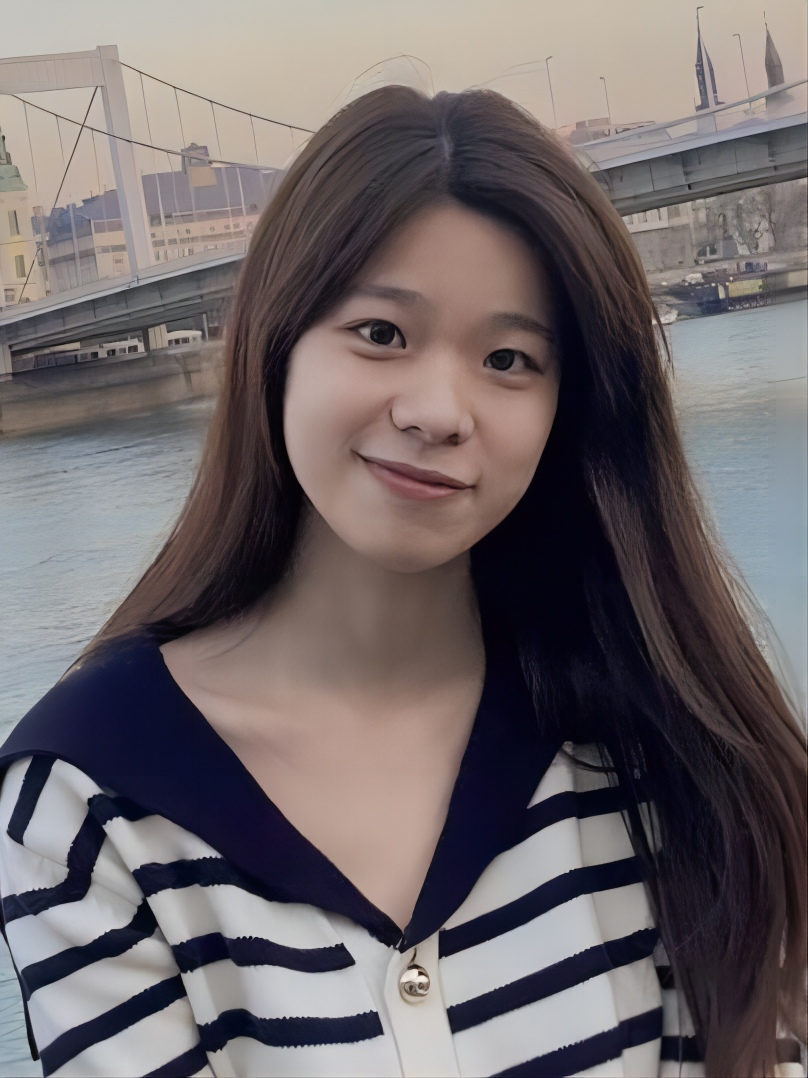}}]{Yuping Yan} received the B.Sc. degree in computer science from University of Shanghai for Science and Technology in 2017 and the double master's degrees in cyber security from the University of Trento (Italy) and E\"otv\"os Lor\'and University (Hungary) in 2017 and 2018, respectively. She is currently a Ph.D. candidate in the faculty of algebra, E\"otv\"os Lor\'and University, Hungary. Her research spans identity management, authentication, cloud computing, and federated learning. Currently, she focuses on attribute-based encryption, formal protocol verification, and privacy-preserving federated learning in medical health applications.
\end{IEEEbiography}

\begin{IEEEbiography}
   [{\includegraphics[width=1in,height=1.25in,clip,keepaspectratio]{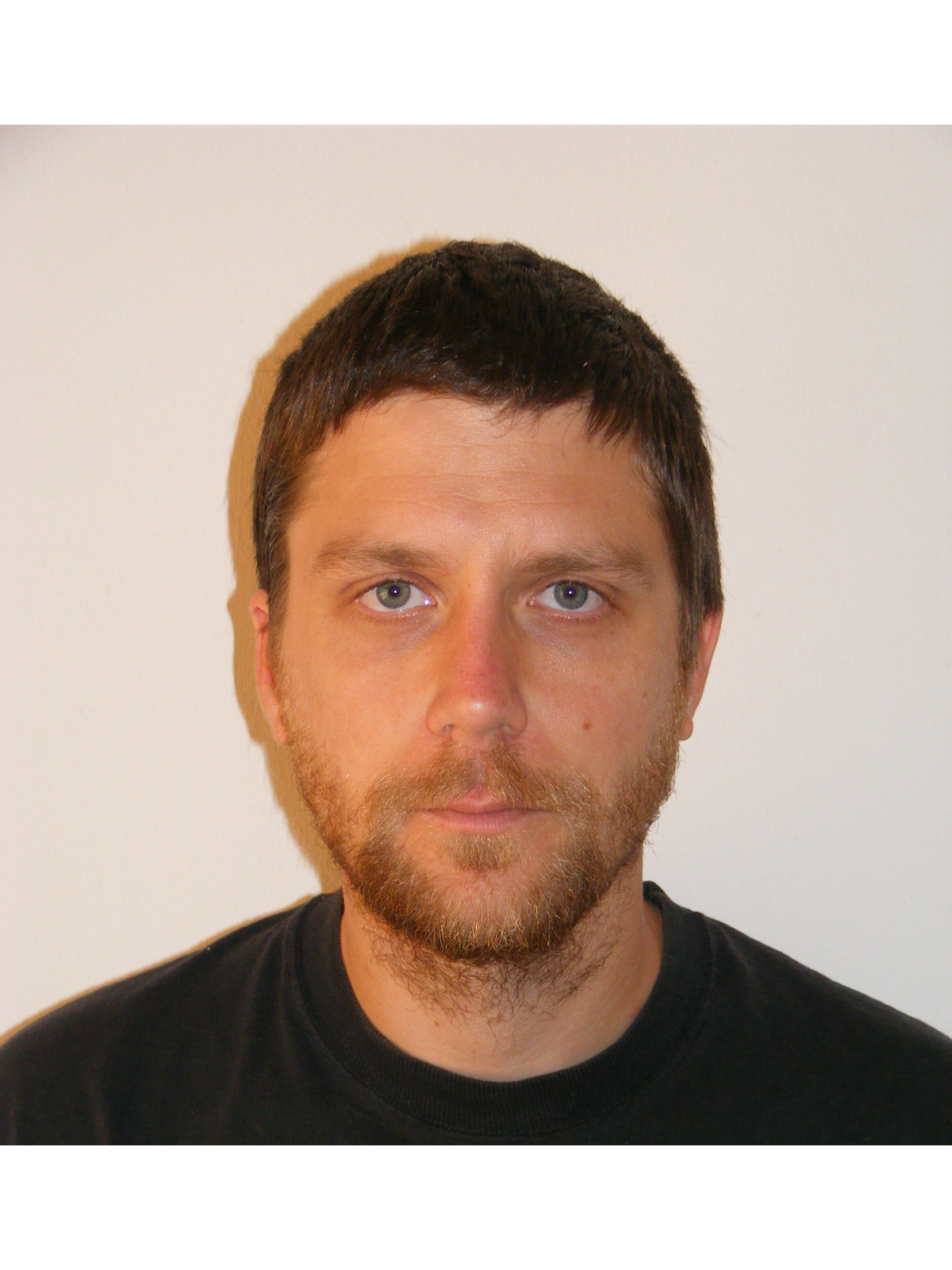}}]{P\'eter Ligeti} received the Ph.D. degree in mathematics and computer science in 2008 from E\"otv\"os Lor\'and University where he is a habilitated associate professor at the Department of Computeralgebra on Faculty of Informatics. He has authored more than 50 research papers and participated in more than 10 research projects. His main research interests cover various fields of combinatorics and cryptography, especially combinatorial optimization, secret sharing and secure distributed communication protocols.
\end{IEEEbiography}

\begin{IEEEbiography}
    [{\includegraphics[width=1in,height=1.25in,clip,keepaspectratio]{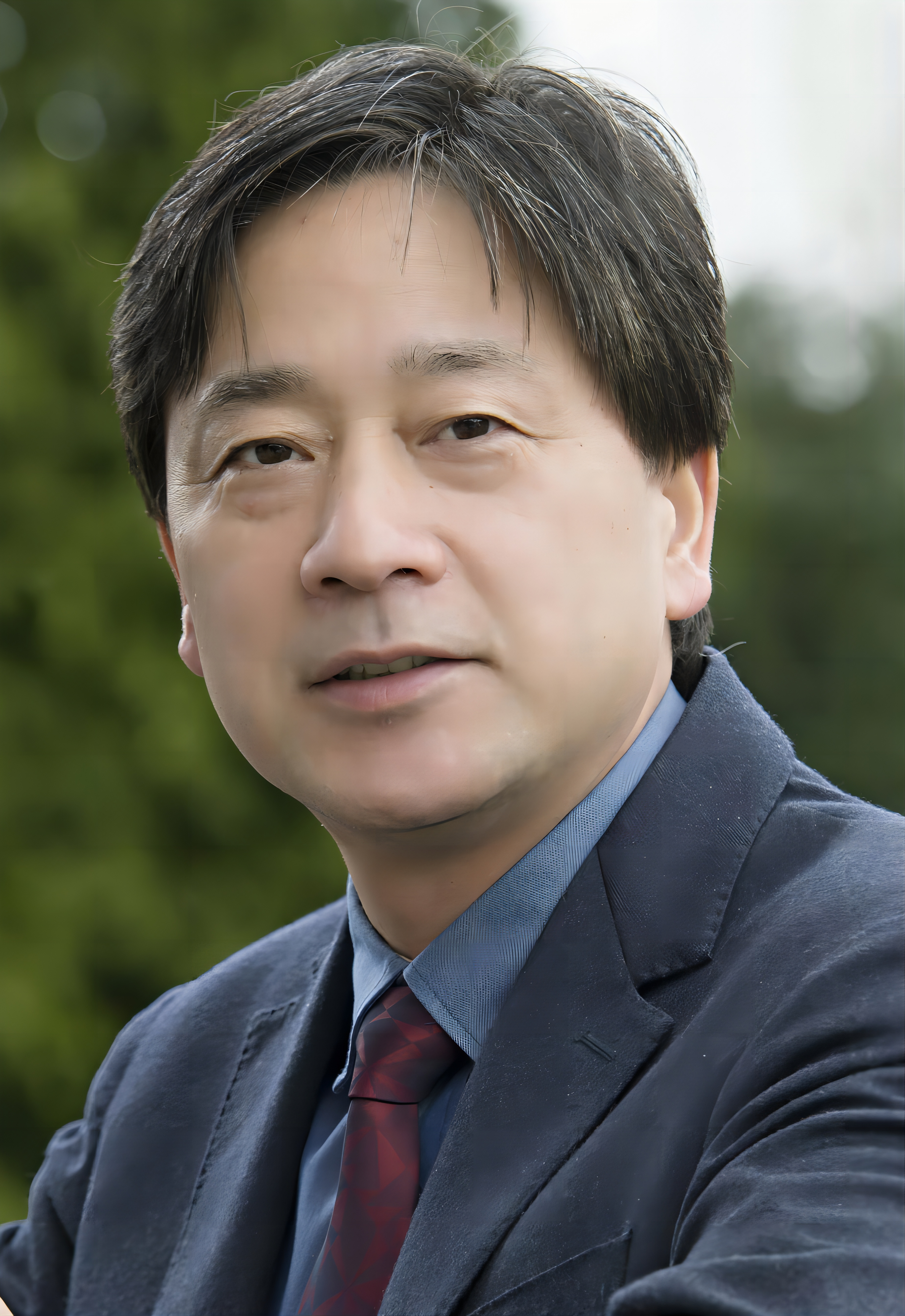}}]{Yaochu Jin} (Fellow, IEEE) received the B.Sc., M.Sc., and Ph.D. degrees in automatic control from Zhejiang University, Hangzhou, China, in 1988, 1991, and 1996, respectively, and the Dr.-Ing. degree from Ruhr-University Bochum, Bochum, Germany, in 2001.

He is an Alexander von Humboldt Professor for Artificial Intelligence endowed by the German Federal Ministry of Education and Research, Chair of Nature Inspired Computing and Engineering, Faculty of Technology, Bielefeld University, Germany. He is also a Distinguished Chair, Professor in Computational Intelligence, Department of Computer Science, University of Surrey, Guildford, U.K. He was a “Finland Distinguished Professor” of University of Jyv\"askyl\"a, Finland, “Changjiang Distinguished Visiting Professor”, Northeastern University, China, and “Distinguished Visiting Scholar”, University of Technology Sydney, Australia. His main research interests include evolutionary optimization, evolutionary learning, trustworthy machine learning, and evolutionary developmental systems. 

Prof Jin is currently the Editor-in-Chief of Complex \& Intelligent Systems. He is the recipient of the 2018, 2021 and 2023 IEEE Transactions on Evolutionary Computation Outstanding Paper Award, and the 2015, 2017, and 2020 IEEE Computational Intelligence Magazine Outstanding Paper Award. He was named a "Highly Cited Researcher" by Clarivate from 2019 to 2022 consecutively. He is a Member of Academia Europaea.
\end{IEEEbiography}

\end{document}